\documentclass{article}

\usepackage[final]{neurips_2024}

\usepackage{mathtools}
\usepackage[utf8]{inputenc} 
\usepackage[T1]{fontenc}    
\usepackage{hyperref}       
\usepackage{url}            
\usepackage{booktabs}       
\usepackage{amsfonts}       
\usepackage{nicefrac}       
\usepackage{microtype}      
\usepackage{xcolor}         

\usepackage{subfigure}
\usepackage{multirow}
\usepackage{varwidth}
\usepackage{pifont}
\usepackage{makecell}
\usepackage{wrapfig}
\usepackage{multicol}
\usepackage{graphicx}
\usepackage{grffile} 
\usepackage{color}
\usepackage{colortbl,booktabs}
\usepackage{subfigure}
\usepackage{parskip}
\usepackage{ulem}

\usepackage{amsmath}
\usepackage{amsfonts}
\usepackage{algorithm}
\usepackage{algpseudocode}

\definecolor{pli-color}{HTML}{002FA7}
\definecolor{yx-color}{HTML}{87CEEB}

\definecolor{xy-color}{HTML}{F0A30A}

\definecolor{gh-color}{HTML}{00FF00}

\definecolor{c-blue}{HTML}{4B9CD3}
\definecolor{c-navy}{HTML}{13294B}
\definecolor{c-pink}{HTML}{EF426F}
\definecolor{c-teal}{HTML}{00A5AD}
\definecolor{c-yellow}{HTML}{FFD100}
\definecolor{c-green}{HTML}{C4D600}

\newcommand{\whichSixteen}{\textcolor{c-blue}{\ding{71}}}
\newcommand{\whichTwelve}{\textcolor{c-navy}{\ding{71}}}
\newcommand{\whichEight}{\textcolor{c-pink}{\ding{71}}}
\newcommand{\whichFourChat}{\textcolor{c-teal}{\ding{71}}}
\newcommand{\whichFourDomain}{\textcolor{c-yellow}{\ding{71}}}
\newcommand{\whichTwo}{\textcolor{c-green}{\ding{71}}}
\newcommand{\whichNone}{\textcolor{white}{\ding{71}}}

\title{\texttt{Model-GLUE}: Democratized LLM Scaling for A Large Model Zoo in the Wild}

\author{
{\normalfont Xinyu Zhao$^{*1}$, ~ Guoheng Sun$^{*2}$, ~  Ruisi Cai$^{*3}$, ~ Yukun Zhou$^{*4}$, ~Pingzhi Li$^{*1}$, ~ Peihao Wang$^{*3}$} \\[0mm]
{Bowen Tan$^{5}$,~ Yexiao He$^{2}$,~  Li Chen$^{6}$,~ Yi Liang$^{6}$,~ Beidi Chen$^{5}$,~ Binhang Yuan$^{4}$}\\[0mm]
{Hongyi Wang$^{\dag7}$,~  Ang Li$^{\dag2}$,~ Zhangyang Wang$^{\dag3}$,~  Tianlong Chen$^{\dag1}$}\\[2mm]
$^{1}$UNC CH ~ $^{2}$UMD ~ $^{3}$UT Austin ~ $^{4}$HKUST ~ $^{5}$CMU ~ $^{6}$Google ~$^{7}$Rutgers University\\[2mm]
\texttt{\{xinyu,pingzhi,tianlong\}@cs.unc.edu, \{ghsun,yexiaohe,angliece\}@umd.edu} \\
\texttt{\{ruisi.cai,peihaowang,atlaswang\}@utexas.edu} \\
\texttt{yzhoufw@connect.ust.hk, \{btan2,beidic\}@andrew.cmu.edu} \\
\texttt{li.lizliz.chen@gmail.com, yiliang@google.com} \\
\texttt{biyuan@ust.hk, hongyi.wang.001@rutgers.edu} \\
\thefootnote{$^{*}$Equal Contribution $^{\dag}$Equal Supervision}
}

\begin{document}


\maketitle

\begin{abstract}
As Large Language Models (LLMs) excel across tasks and specialized domains, scaling LLMs based on existing models has gained significant attention, which is challenged by potential performance drop when combining disparate models. 
Various techniques have been proposed to aggregate pre-trained LLMs, including model merging, Mixture-of-Experts, and stacking. Despite their merits, a comprehensive comparison and synergistic application of them to a diverse model zoo is yet to be adequately addressed.
In light of this research gap, this paper introduces \texttt{Model-GLUE}, a holistic LLM scaling guideline. 
First, our work starts with a benchmarking of existing LLM scaling techniques, especially selective merging, and variants of mixture. 
Utilizing the insights from the benchmark results, we formulate a strategy for the selection and aggregation of a heterogeneous model zoo characterizing different architectures and initialization.
Our methodology involves clustering mergeable models, selecting a merging strategy, and integrating model clusters through model-level mixture. Finally, evidenced by our experiments on a diverse Llama-2-based model zoo, \texttt{Model-GLUE} shows an average performance enhancement of 5.61\%, achieved without additional training.
Codes are available at \url{https://github.com/Model-GLUE/Model-GLUE}.

\end{abstract}

\section{Introduction}\label{sec:introduction}

\vspace{-2mm}
Large Language Models (LLMs) have demonstrated unparalleled capability in a diverse array of natural language tasks, encompassing commonsense reasoning, question answering, and specialized domains such as mathematics and programming~\cite{Achiam2023GPT4TR,Rozire2023CodeLO,Touvron2023Llama2O}. The effectiveness of LLMs is based on the scaling law, which posits that proportionally increasing model and training data size leads to enhanced model performance~\cite{Kaplan2020ScalingLF}. Nevertheless, the computation overhead and data requirement surge as LLM continues to scale. With the widespread of open-sourced general or specialized LLMs, aggregating existing models to construct a more versatile LLM emerges as an economical alternative to training a larger LLM from scratch  ~\cite{ding2024mastering,goddard2024arcee,wan2024knowledge}. This not only mitigates the computation cost but also leverages the collective advancements of previous efforts in building LLMs. 

\vspace{-2mm}
Within different methods to combine existing LLMs, a major class is merging~\cite{ainsworth2022git,akiba2024evolutionary,ilharco2023editing,jang2024model,matena2022merging,wortsman2022model,yadav2023tiesmerging,yu2024language}. Model merging combines multiple models into a single one of the same size through weight-space transformation.~\citet{wortsman2022model} first propose to merge a few fine-tuned models as a training trick for the flat loss-landscape, and~\citet{ilharco2023editing} extends it to multi-task scenario, both of which employ the simple averaging.
Other works propose more complicated merging methods, leveraging weight sparsity~\cite{yadav2023tiesmerging, yu2024language} and non-uniform coefficient~\cite{akiba2024evolutionary,matena2022merging}. However, they assume that all candidate models are ``useful'' when merging. While this may hold for small-sized designed model collections, it may not be the case in real-world scenarios given a large and divergent model zoo. 
How to ensure the benefits of merging different model zoo sizes and similarities, and exclude ``harmful'' candidates, remains underexplored.

\vspace{-2mm}
Since merging is limited to the same model structures and initial weights, another alternative is Mixture-of-Experts (MoE)~\cite{goddard2024arcee}. MoE is a conditional computation architecture that activates only a subset of model parameters for each specific input example~\cite{shazeer2017outrageously}. MoE LLMs have already demonstrated performance and computational efficiency advantages over their dense counterparts~\cite{switch,jiang2024mixtral,gshard,Zoph2022DesigningES}. In particular, we use a broader term ``mixture'' to denote the aggregation of existing expert LLMs according to the MoE paradigm, which has been successfully implemented in some recent practices~\cite{sukhbaatar2024branchtrainmix,wan2024knowledge,wang2023fusing}. However, these implementations neglect the inherent flexibility of MoE to integrate different expert models, especially those groups that do not work with merging. Also, the difference and possible synergy between merging and mixing have not been thoroughly investigated. Based on the above challenges, our primary research question is formulated as:

\vspace{-1mm}
\textit{(Q) Is it feasible to establish a benchmark for selecting and aggregating Large Language Models (LLMs) from an extensive and varied model zoo based on current state-of-the-art model merging and mixture, thereby enhancing the overall competence of the final model?}

\begin{wrapfigure}{}{0.4\textwidth}
\vspace{-6mm}
\resizebox{\linewidth}{!}{
\includegraphics[]{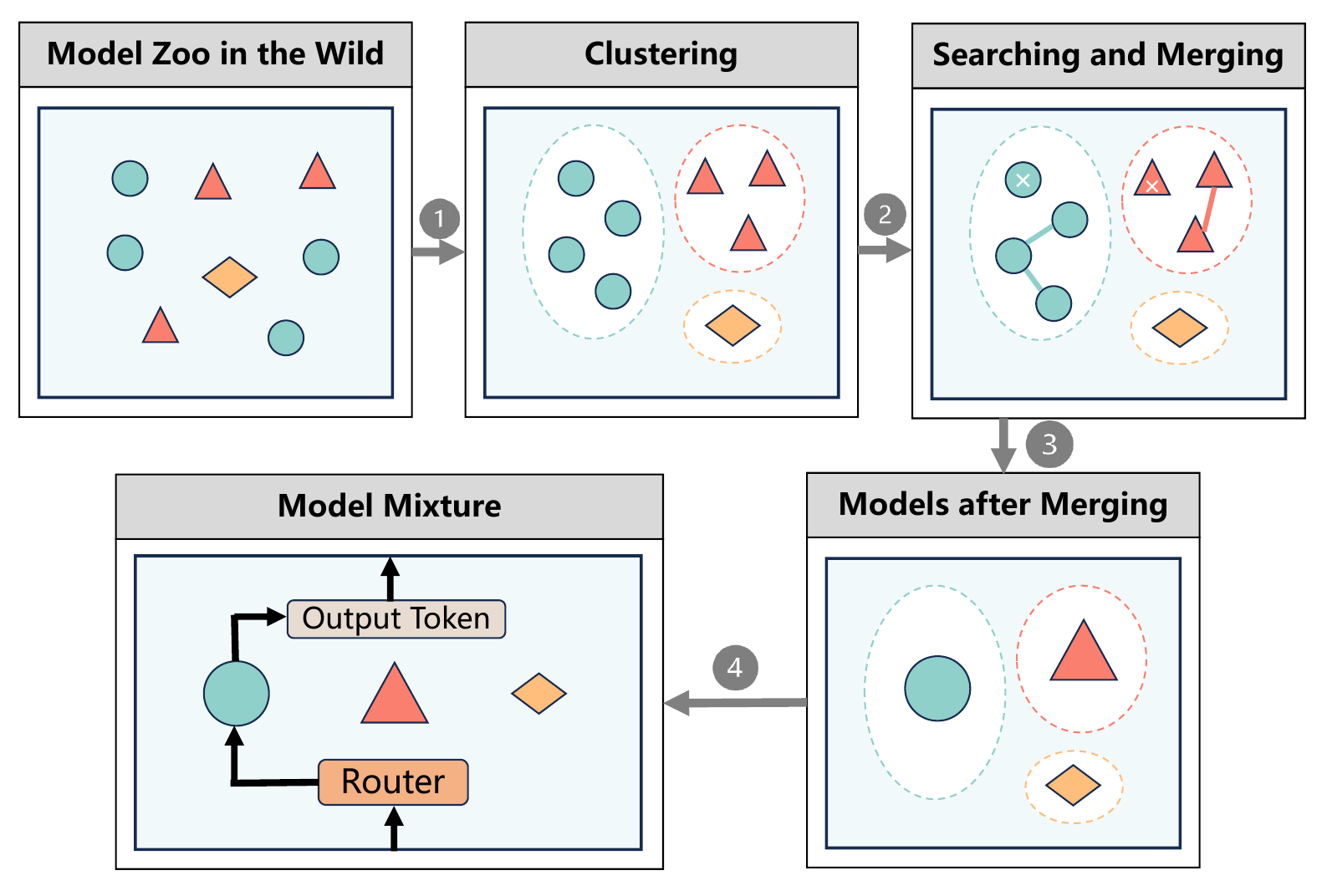}}
\vspace{-8mm}
\caption{\small Overview of \texttt{Model-GLUE}, composing of (1) Model Clustering based on architecture and weight similarity; (2) Model Filtering and Searching for merging; (3) Model Merging within each cluster; (4) Model Level Mixture of merged models.}
\label{lmm}
\vspace{-5mm}
\end{wrapfigure}

\vspace{-1mm}
To address (Q), we present \texttt{Model-GLUE}, a comprehensive benchmark and set of guidelines for LLM scaling. \texttt{Model-GLUE} is the first work for LLM scaling encompassing a wide range of model group sizes and variability, with a principal emphasis on the merging and mixture methodologies, and also discussion of model stacking. We first delve into merging scheduling, analyzing strategies for identifying potentially detrimental model candidates and various merging techniques. We then explore a variety of model mixtures as an alternative to merging, covering different mixture granularity, routers architecture, routing input inputs, \textit{etc.} 
Building upon the insights from model merging and mixture, \texttt{Model-GLUE} introduces an efficient and robust LLM scaling recipe for a diverse set of models. It starts with model clustering and progressive merging, and then the mixture of all clusters, thereby integrating similar knowledge from the model zoo while highlighting the respective strengths of each cluster. Our contributions are outlined as follows:

\vspace{-1mm}
$\bullet$ 
We conduct a comprehensive benchmarking analysis of LLM merging strategies, beginning with identifying each model's contribution and then followed by filtering out detrimental candidates. Our findings are validated on a range of LLMs, from a few to over a dozen.

\vspace{-2mm}
$\bullet$ 
We assess model mixture for four distinct variants: mixture level, router design, router input, and hybrid mixture. We have derived several principles for model mixture and discussed its utility as a solution for scaling models incompatible with merging. 

\vspace{-2mm}
$\bullet$ 
We introduce a recipe for progressively combining LLM models, \texttt{Model-GLUE}, based on findings on merging and mixture benchmarks. It first conducts selective merging and then model mixture, outperforming the best single model on general reasoning, mathematics, and coding tasks.

\vspace{-2mm}
$\bullet$ Extensive experimental results on Llama-2-based models validate our proposal. For instance, \texttt{Model-GLUE} achieves an average increase of $5.61\%$ across chatting, mathematics, and coding benchmarks compared to the best single LLM.
\section{Related Works}

\paragraph{Model Merging.} 
Merging methods can be divided into zero-shot merging and merge-then-train approaches. Early zero-shot merging methods are weight averaging and Linear Mode Connectivity~\cite{nagarajan2021uniform,wortsman2022model}. Later popular methods include Task Arithmetic~\cite{ilharco2023editing} manipulating task vectors, and TIES~\cite{yadav2023tiesmerging} addressing parameter interference through trimming and conflict resolution. DARE~\cite{yu2024language} optimizes parameters selectively to enhance merging without extra training. Others focus on geometric properties of weights for merging~\cite{10.1145/325334.325242,jang2024model}. Recent Evolutionary Model Merge~\cite{akiba2024evolutionary} improves weight configuration and data token pathways during inference. 
For the merge-then-train approach, Fisher merging~\cite{matena2022merging} uses the Fisher information matrix to weigh model parameters to maximize their joint likelihood. RegMean~\cite{jin2023dataless} adapts the linear merging to each linear layer while averaging embeddings and biases.
However, both zero-shot and merge-then-train approaches are less effective for models initialized differently.~\cite{ainsworth2022git,imfeld2023transformer,verma2024merging,xu2024training} exploit the permutation symmetry inherent in neural networks on small to large models. 
To boost merging efficiency, our focus on merging lies in the zero-shot merging of models with the same architecture and initialization.

\vspace{-2mm}
\paragraph{Model Mixture.} 
Mixture-of-Experts~(MoE)~\cite{shazeer2017outrageously} scales up neural networks by utilizing router networks to activate different parts of the model for different input tokens. Its integration with Large Language Models (LLMs) has gained notable recognition for its exceptional generative capabilities and unparalleled efficiency. Recently, Mixtral~\cite{jiang2024mixtral} demonstrates that the MoE methodology can achieve the performance of dense LLM counterparts while employing significantly fewer active parameters. Model mixture combines a collection of dense LLM models, irrespective of their sizes, into a MoE model. Some studies discover model fusion \cite{wan2024knowledge,wang2023fusing} integrating the outputs of expert models to exploit the unique insights into the data distribution. Recent initiatives include Branch-Train-MiX~\cite{sukhbaatar2024branchtrainmix}, which starts with a seed-dense LLM and then branches out, facilitating the parallel training of expert models. These trained dense models are subsequently incorporated as experts within MoE layers, with other parameters being averaged. However, this approach is limited to dense models that share identical architectures and sizes. Most recently, UltraFuser~\cite{ding2024mastering} introduces a token-level soft gating mechanism that blends model outputs, with a two-stage training strategy.

\vspace{-2mm}
\paragraph{Model Stacking.}

Model stacking concatenates two models along the depth dimension.
In the era of LLM,~\citet{wu2024llama} reuses pre-trained LLaMA layers and resets the output projection to zero in stacking.~\citet{kim2023solar} shows dropping middle layers in stacking yields superior performance.~\citet{wang2023data} prove that stacking could help recover model-parameter scaling laws with insufficient data. \citet{reddi2023efficient} demonstrated that gradual stacking leads to significant improvements in wall-clock time during the training of few-shot learners. Theoretically, \citet{agarwal2024stacking} proved that model stacking could be interpreted as Nesterov acceleration in network optimization. However, all the aforementioned stacking methods involve no more than two kinds of models and primarily focus on the benefits of training acceleration.
In this work, we explore the possibility of stacking two heterogeneous models to combine their capabilities.

\vspace{-2mm}
\paragraph{Model Scaling Tools} 
There have been several tools for model mixture and merging, and for scaling models using existing LLMs. For example, Mergekit is an open-source library designed to facilitate the application of model merging strategies and the construction of MoE~\cite{goddard2024arcee}. As a representative of unified LLM, Beyonder is a set of mixtures of merged and single LLMs for different tasks\footnote{\url{https://huggingface.co/mlabonne/Beyonder-4x7B-v3}}. However, there is still a lack of a comprehensive benchmark of the various mixing and merging techniques and practical guidance on how to unify groups of LLMs at different levels of similarity. 

\vspace{-1mm}
\section{Methodology}
\vspace{-1mm}
\subsection{Preliminaries} 

In this study, we consider a collection of $n$ existing Large Language Models (LLMs), denoted as $\{ \mathtt{M}_1, \ldots, \mathtt{M}_n \}$, which have been fine-tuned on diverse corpora. Our objective is to outline a systematic approach towards producing one stronger aggregated model across all knowledge domains. Specifically, the unified LLM incorporates single LLMs mainly through merging and mixture.
\vspace{-1mm}
\subsection{Model Merging}\label{sec:merging_method}

\begin{wrapfigure}{r}{0.5\linewidth}
\centering
\vspace{-4mm}
\includegraphics[width=1\linewidth]{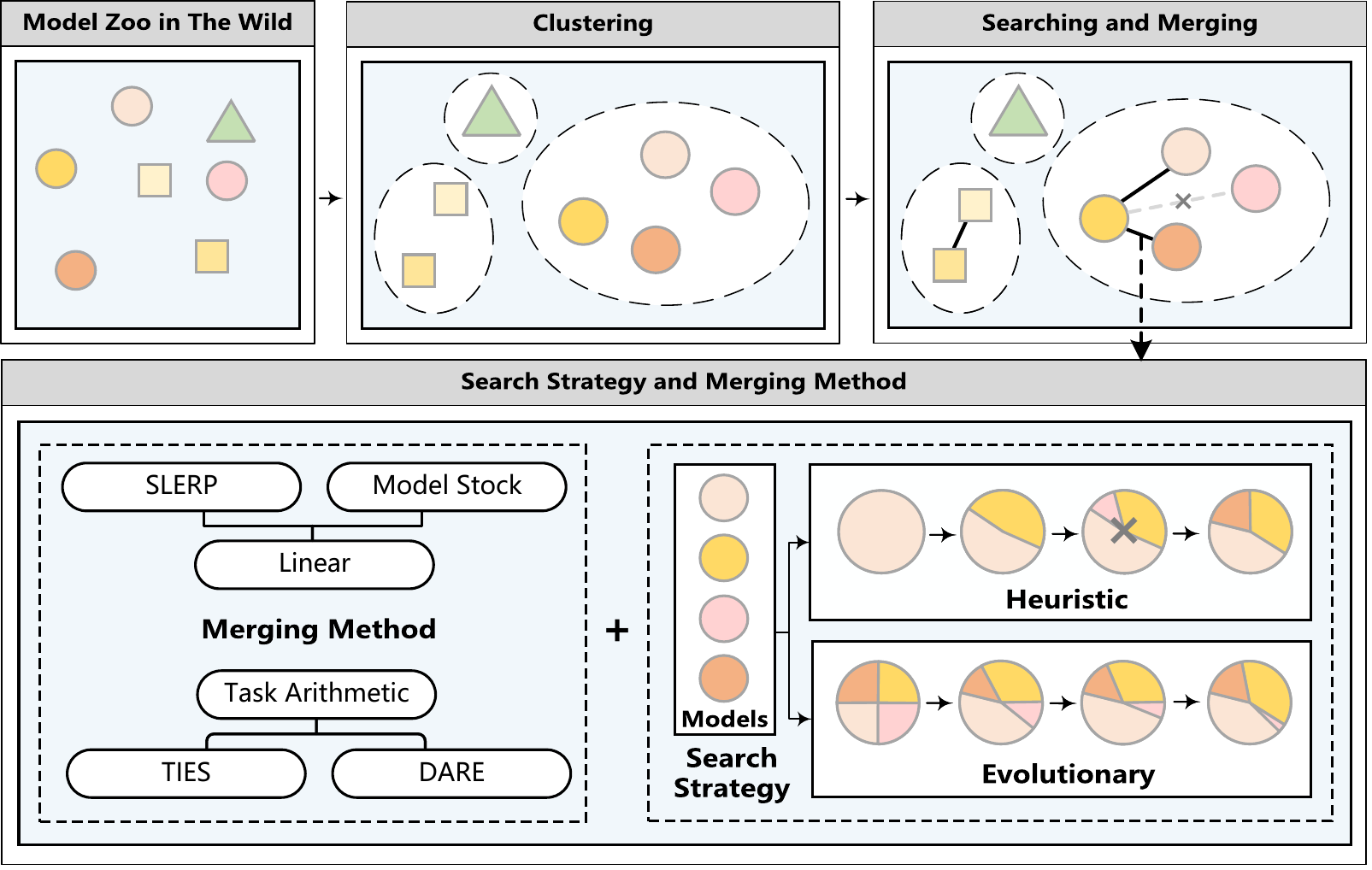}
\centering
\vspace{-6mm}
\caption{\small Pipeline for model merging, as well as an overview of merging methods and search strategies.}
\vspace{-6mm}
\label{merging_pipline}
\end{wrapfigure}
\paragraph{The concept of Model Merging} 

Model merging is integrating multiple models into one unified model in the weight space, compatible with LLMs of the same initialization~\cite{goddard2024arcee}.  
Popular merging methods can be divided into two types: \ding{182}
\textit{Merging entire model weights} represented by Model Soup~\cite{wortsman2022model} (Linear), SLERP~\cite{10.1145/325334.325242}, and Model Stock~\cite{jang2024model}; \ding{183} \textit{Task-vector based merging} represented by Task Arithmetic~\cite{ilharco2023editing}, TIES~\cite{yadav2023tiesmerging},  and DARE~\cite{yu2024language}.
The former method directly interpolates model weights, while the latter subtracts the pre-trained model from the fine-tuned model to obtain task vectors and utilizes sparsity and consistency of parameters for refined merging.
The basic Linear interpolation merging is defined as $w_{u}=\sum_{i=1}^{n}s_{i}\cdot w_{i}$, where $w_{i}$ and $s_{i}$ are the corresponding model weights and merging coefficient of $\mathtt{M}_i\in\{\mathtt{M}_1,\ldots \mathtt{M}_n\}$.

\vspace{-2mm}
\paragraph{Selective Merging Pipeline} Merging can be easily applied to models with the same architecture, but does not guarantee better results. 
Therefore, before searching for the merging coefficient, we first pre-process the models by clustering all the models using cosine similarity and then searching for the optimal merging coefficient and method within each cluster. Details are explained in Appendix~\ref{apdx:clustering}.

\vspace{-2mm}
\paragraph{Heuristic and Evolutionary Strategies} 
The heuristic strategy is for searching and filtering potential harmful models for merging. It is based on greedy search, involving three variants: \ding{182} \textit{Heuristic-Average} 
retain the candidate if there is an improvement on the proxy dataset in each round of merging.
\ding{183} \textit{Heuristic-Coefficient} builds upon \textit{Heuristic-Average}, by combining the previously merged model with a new candidate using different coefficients in each round. \ding{184} \textit{Heuristic-Similarity} selects the candidate model with the highest or lowest similarity and conducts a coefficient search to combine it with the previously merged model. Detailed heuristic strategy algorithms can be found in Appendix~\ref{merge_algo}
Heuristic strategies perform pairwise merging of models, while many methods allow for merging multiple models at once. Therefore, we also consider jointly optimizing all model coefficients using the \textit{Evolutionary Strategy}. 

\subsection{Model Mixture} 
\begin{wrapfigure}{r}{0.6\linewidth}
\centering
\vspace{-5mm}
\includegraphics[width=1\linewidth]{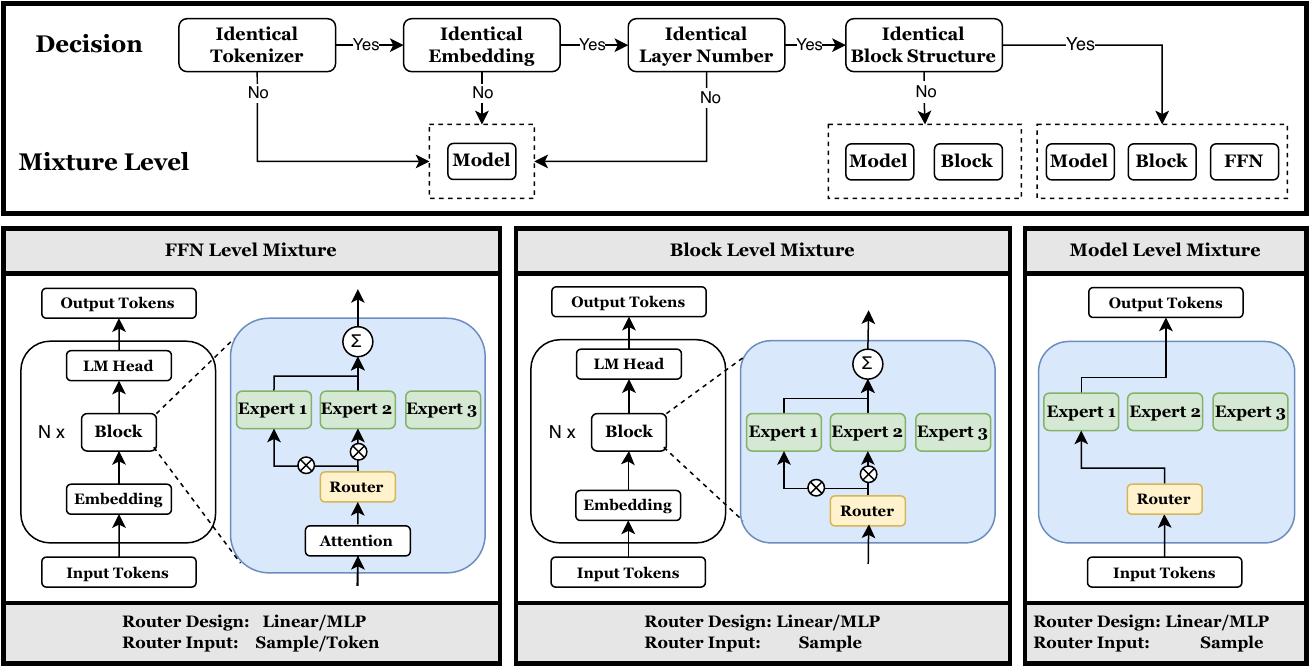}
\centering
\vspace{-4mm}
\caption{\small The overview and decision flow of three model mixture levels and their selection philosophy. }
\vspace{-4mm}
\label{model_mixture}
\end{wrapfigure}

\paragraph{The concept of Model Mixture.} Model mixture resembles Mixture-of-Experts(MoE). It scales a LLM with multiple pre-trained LLM experts and further extends beyond traditional token-dependent Feed-Forward-Network (FFN) MoE designs~\cite{shazeer2017outrageously}. A mixture model is composed of MoE modules and the rest shared parameters. A MoE module consists of a router $\mathcal{G}(\cdot)$ and $n$ expert networks $\{\mathtt{E}_1,\cdots,\mathtt{E}_{n}\}$. $\mathcal{G}(\cdot)$ takes a router input $\boldsymbol{x_\mathcal{G}}$ and generate expert assignment for each token input $\boldsymbol{x}$. Then MoE outputs a weighted sum of experts' outputs as $\mathtt{MoE}(x,x_\mathcal{G}) = \sum_{i=1}^{n}\mathcal{G}(x_\mathcal{G})_{i}\cdot\mathtt{E}_i(x)$. We experiment with several variations of Model Mixture, classified as follows:

\vspace{-1mm}
\paragraph{Mixture levels.} 
Traditional Mixture-of-expert models replace the dense FFN layer at each Transformer block with an MoE module, which is only compatible with LLMs that share the same architecture. Besides this \ding{182} \textit{FFN level mixture}, we also experiment with two coarse-grained mixtures. \ding{183} \textit{Block level mixture} create MoE module by aggregating Transformer blocks with the same index from each LLM as experts and add a block-wise router. Block level mixture is applicable to models with different architecture but the same embedding space, layer amounts, and intermediate dimension. \ding{184} \textit{Model level mixture} take each LLM as an expert and use a router at mixture model input. Model level mixture covers any LLM groups not compatible with FFN and block level mixture. In particular, the model level mixture is similar but not identical to the model ensemble, as the former can be sparse and focus more on efficiency and exploit single LLM expertise, while the latter produces general results by averaging or majority voting overall model outputs. Details can be found in Appendix~\ref{apdx:algo_mixture}

\vspace{-1mm}
\paragraph{Router design.} 
The router network of many MoE studies adheres to a \ding{182} \textit{linear router}~\cite{shazeer2017outrageously}. We experiment with another more complex \ding{183} \textit{MLP router} to examine whether this router design leads to better performance. It is implemented by two sequential FFN and a ReLU function in between, inspired by~\cite{Shen2023ModuleFormerLM,Liang2022M3ViTMV}. For the routing method, we employ Top-K selection to all routers, which activates the K experts corresponding to the K largest softmaxed router output~\cite{shazeer2017outrageously,Shen2023ModuleFormerLM}.

\vspace{-1mm}
\paragraph{Router input.} 
We adopt two types of router input for different levels of model mixture: \ding{182} Token input for FFN level mixture, where router input is the same as model input; \ding{183} Sample input for block and model level mixture, where we calculate the average embedding as the sample input $x_\mathcal{G} = \sum_{i=1}^{n}x_n$, and route tokens of a sample to the same expert based on sample routing. The sample routing avoids inconsistency in attention operation. 

\vspace{-1mm}
\paragraph{Hybrid mixture.} 
To explore LLM scaling in between model merging and model mixture, we propose the hybrid mixture as an intermediate solution. In a hybrid mixture, the bottom few layers of all single LLMs are merged, and then the rest layers follow any of the mixture level designs.


\section{Model Merging and Model Mixture for LLMs}\label{Benchmarks}
\subsection{Benchmark Datasets and Configs}
\label{sec:exp:configs}
\vspace{-1mm}
\paragraph{Model Zoo.}
Table~\ref{tab:model-zoo} provides an overview of the Model Zoo.
For benchmarking model merging and mixture at different sizes of model zoo, we construct $5$ groups of Llama-2-based $7$B chat LLMs where the number of models $\in [ 2, 4, 8, 12, 16]$. In addition, to examine the difference in combining models from different domains, we introduce \texttt{Which4~(chat)}, consisting of four chat models, as a supplement setting where no single model has a superior advantage in a specific domain. 

\vspace{-2mm}
After comparing the two ways of model scaling, we propose \texttt{Model-GLUE} combining selective merging and model mixture, which is tested on the largest model family \texttt{Which16}. 
\texttt{Which16} is developed on $12$ mergeable Llama-2-based models in \texttt{Which12}, which additionally includes four highly performant domain-specific models that cannot be merged: three CodeLlama-based models, two of which are code models and one is a math model, and LLM360/CrystalChat.
In particular, LLM360/CrystalChat use different architecture, initialization, and training data from Llama-2-based models, while CodeLlama series, initialized from Llama-2, adopt continuous pretraining rather than fine-tuning as models in \texttt{Which12}.

\vspace{-3mm}
\begin{table}[ht]
\centering
\caption{\small All of the models in our model zoos and their performance. For each model zoo, we denote those models that belong to it with a colored star \ding{71}: \whichTwo~for \texttt{Which2}, \whichFourChat~for \texttt{Which4~(Chat)}, \whichFourDomain~for \texttt{Which4~(Domain)}, \whichEight~for \texttt{Which8}, \whichTwelve~for \texttt{Which12}, and \whichSixteen~for \texttt{Which16}.} \label{tab:model-zoo}
\renewcommand\arraystretch{1}
\resizebox{1\linewidth}{!}{
\begin{tabular}{l|r|ccccccc}
\toprule
\midrule
Model & Model Zoo & ARC  & WinoGrande  & MMLU & GSM8K & MBPP & HumanEval & Average \\ 
\midrule
migtissera/Synthia-7B-v1.2~\cite{mukherjee2023orca,Synthia-7B-v1.2,Touvron2023Llama2O} & \whichFourChat \whichFourDomain \whichEight \whichTwelve \whichSixteen & $55.03\%$ & $73.72\%$ & $48.18\%$ & $24.03\%$ & $17.80\%$ & $13.41\%$ & $38.70\%$ \\
neuralmagic/Llama-2-7b-evolcodealpaca~\cite{Touvron2023Llama2O} & \whichFourDomain \whichEight \whichTwelve \whichSixteen & $49.57\%$ & $72.45\%$ & $41.70\%$ & $09.02\%$  & $25.60\%$ & $31.71\%$ & $38.34\%$ \\
teknium/OpenHermes-7B~\cite{Touvron2023Llama2O} & \whichFourChat \whichNone \whichEight \whichTwelve \whichSixteen & $56.40\%$ & $73.88\%$ & $47.84\%$ & $09.25\%$  & $22.80\%$ & $19.51\%$ & $38.28\%$ \\
PygmalionAI/pygmalion-2-7b~\cite{Touvron2023Llama2O} & \whichFourDomain \whichEight \whichTwelve \whichSixteen & $54.10\%$ & $75.37\%$ & $48.38\%$ & $17.29\%$ & $19.20\%$ & $15.24\%$ & $38.26\%$ \\
meta-llama/Llama-2-7b-chat-hf~\cite{Touvron2023Llama2O} & \whichTwo \whichFourChat \whichNone \whichEight \whichTwelve \whichSixteen & $54.10\%$ & $71.27\%$ & $47.28\%$ & $23.05\%$ & $17.00\%$ & $13.41\%$ & $37.68\%$ \\
Severus27/BeingWell\_llama2\_7b~\cite{Touvron2023Llama2O} & \whichTwelve \whichSixteen & $54.95\%$ & $72.30\%$ & $46.19\%$ & $22.29\%$ & $13.40\%$ & $13.41\%$ & $37.09\%$ \\
meta-math/MetaMath-7B-V1.0~\cite{Touvron2023Llama2O,yu2023metamath} & \whichFourDomain \whichEight \whichTwelve \whichSixteen & $47.35\%$ & $70.24\%$ & $41.58\%$ & $59.06\%$ & $01.40\%$  & $01.22\%$  & $36.81\%$ \\
lmsys/vicuna-7b-v1.5~\cite{zheng2023judging,Touvron2023Llama2O} & \whichTwo \whichFourChat \whichNone \whichEight \whichTwelve \whichSixteen & $53.75\%$ & $70.56\%$ & $49.78\%$ & $19.11\%$ & $06.00\%$  & $19.51\%$ & $36.45\%$ \\
garage-bAInd/Platypus2-7B~\cite{hu2022lora,platypus2023,Touvron2023Llama2O} & \whichEight \whichTwelve \whichSixteen & $55.12\%$ & $74.03\%$ & $49.82\%$ & $02.50\%$  & $19.00\%$ & $14.63\%$ & $35.85\%$ \\
GOAT-AI/GOAT-7B-Community~\cite{bekbayev2023poison,Touvron2023Llama2O} & \whichTwelve \whichSixteen & $49.06\%$ & $72.22\%$ & $49.23\%$ & $09.70\%$  & $05.40\%$  & $09.76\%$  & $32.56\%$ \\
stanford-oval/Llama-2-7b-WikiChat-fused~\cite{semnani-etal-2023-wikichat,Touvron2023Llama2O} & \whichTwelve \whichSixteen & $50.94\%$ & $68.59\%$ & $39.13\%$ & $00.00\%$  & $13.80\%$ & $04.27\%$  & $29.45\%$ \\
cognitivecomputations/dolphin-llama2-7b~\cite{Touvron2023Llama2O} & \whichTwelve \whichSixteen & $42.66\%$ & $65.35\%$ & $46.52\%$ & $10.69\%$ & $00.80\%$  & $02.44\%$  & $28.08\%$ \\
meta-math/MetaMath-Llemma-7B~\cite{azerbayev2023llemma,yu2023metamath} & \whichSixteen & $46.76\%$ & $64.33\%$ & $46.33\%$ & $62.40\%$ & $42.00\%$ & $31.10\%$ & $48.82\%$ \\
codellama/CodeLlama-7b-Instruct-hf~\cite{rozière2024code} & \whichSixteen & $43.52\%$ & $65.11\%$ & $41.83\%$ & $17.06\%$ & $40.00\%$ & $33.70\%$ & $40.20\%$ \\
ise-uiuc/Magicoder-S-CL-7B~\cite{magicoder,rozière2024code} & \whichSixteen & $43.77\%$ & $63.38\%$ & $35.94\%$ & $14.33\%$ & $50.20\%$ & $63.41\%$ & $45.17\%$ \\ 
LLM360/CrystalChat~\cite{liu2023llm360} & \whichSixteen & $51.54\%$ & $70.64\%$ & $52.39\%$ & $32.45\%$ & $38.80\%$ & $35.37\%$ & $46.87\%$ \\ 
\midrule
\bottomrule
\end{tabular}}
\end{table}
\vspace{-1mm}

For merging benchmarks, we experiment with a larger model zoo, namely \texttt{Which4}, \texttt{Which8}, and \texttt{Which12} with models filtered from \texttt{Which16}.
For model mixture with higher computational cost, we experiment with \texttt{Which2} and \texttt{Which4}.

\vspace{-2mm}
\paragraph{Benchmarks} We assess all models on three categories of benchmarks: (\textit{i}) Commonsense reasoning using ARC~\cite{Clark2018ThinkYH}, WinoGrande~\cite{Sakaguchi2019AnAW}, and MMLU~\cite{Hendrycks2020MeasuringMM}; (\textit{ii}) Mathematics ability on GSM8K~\cite{Cobbe2021TrainingVT}; (\textit{iii}) Coding ability on MBPP~\cite{Austin2021ProgramSW} and HumanEval~\cite{Chen2021EvaluatingLL}. 
The evaluation scripts are based on lm-eval~\footnote{\url{https://github.com/EleutherAI/lm-evaluation-harness}} for commonsense and mathematical reasoning and bigcode-eval~\footnote{\url{https://github.com/bigcode-project/bigcode-evaluation-harness}} for coding datasets. All benchmarks are under the MIT License.

\definecolor{weborange}{RGB}{255,165,0}
\newcommand{\ruisi}[1]{{{\color{weborange}(ruisi) #1}}}

\vspace{-2mm}
\subsection{Implementation Details for Merging} 
\vspace{-1mm}
\paragraph{Proxy Dataset.} Since the performance of merging model is not necessarily positive, we need a proxy dataset to determine whether to reject a particular round of merging in the Heuristic Strategy, or to compute the model fitness in the Evolutionary Strategy.  
 (\textit{i})~For MBPP, we select its validation set. (\textit{ii})~For HumanEval, due to the unavailability of a validation set and its smaller size, we select $20\%$ of the JavaScript version of HumanEvalPack~\cite{muennighoff2023octopack}. (\textit{iii})~For other tasks, we chose the small-scale datasets released by tinybenchmarks~\cite{polo2024tinybenchmarks} under MIT License. 

\vspace{-2mm}
\paragraph{Model Zoo and Clustering.} The Merging Bench considers $3$ model zoos: \texttt{Which4}, \texttt{Which8}, and \texttt{Which16}. We first cluster the model zoos based on cosine similarity with a threshold of $0.95$. Due to \texttt{Which16} contains models that cannot be merged, we choose the mergable family obtained through clustering which is referred to as \texttt{Which12}.

\vspace{-2mm}
\paragraph{Details of Heuristic Strategy and Evolutionary Strategy.} 
For \textit{Heuristic Strategy}, to reduce the search space, we only evaluated Linear interpolation and the range of coefficient search is $\{0.1,0.2...0.9\}$. In \textit{Heuristic-Similarity}, we use the average similarity of all weights as the criterion for selecting models in each round.
For \textit{Evolutionary Strategy}, we refer to the setting of Evolutionary Model Merge~\cite{akiba2024evolutionary}, which utilizes the CMA-ES~\cite{hansen2006cma} algorithm implemented by Optuna~\cite{akiba2019optuna}. In contrast, all parameters are randomly initialized, and the fitness values are defined as the accuracy of the proxy dataset. The optimization was conducted for 200 trials in all scenarios.

\vspace{-2mm}
\subsection{Model Merging Benchmark Results} \label{Merging_Bench}

\begin{figure}[t]
    \centering
    \includegraphics[width=\linewidth]{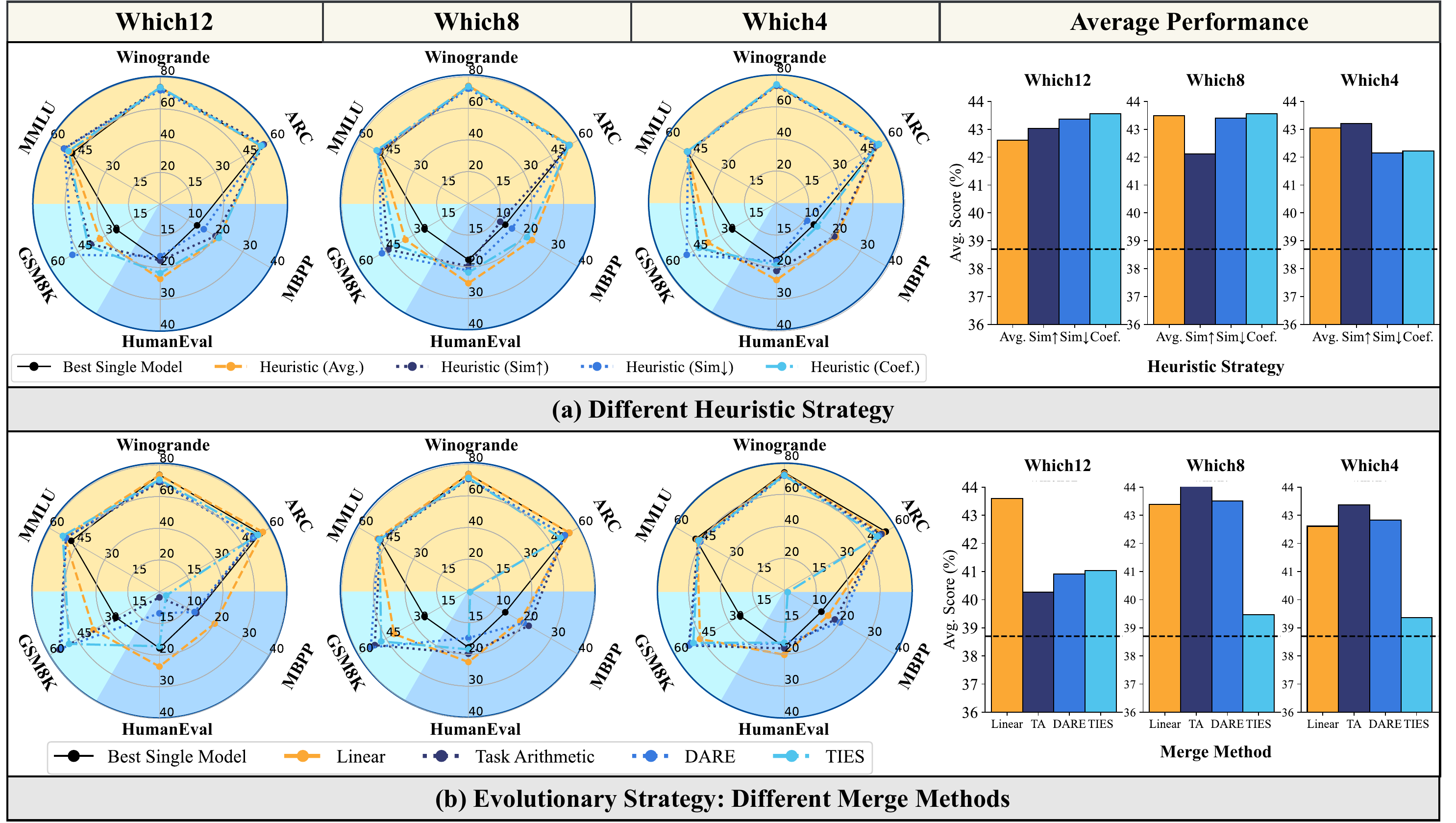}
    \vspace{-6mm}
    \caption{\small (a)~Comparison between different Heuristic Strategies on \texttt{Which12}, \texttt{Which8}, \texttt{Which4}. (b)~Comparison of different model merging methods in Evolutionary Strategy.} \label{fig:merge_radar}
    \vspace{-4mm}
\end{figure}

We start our discussion by examining the effectiveness of existing approaches in depth. Despite existing merging methods focus on improving the merging techniques, their effectiveness is usually validated basedt on small-scale model zoos. For instance, \citet{ilharco2023editing} primarily focuses on the linear interpolation between two fine-tuned models, while \citet{akiba2024evolutionary} explores merging three. 

\vspace{-2mm}
Current model practitioners typically download pre-trained models, fine-tune them on their own data or with unique techniques for specific downstream tasks, and then upload them back to the public. This practice results in a large number of open-source models being available, yet they remain underutilized by current merging methods.
To this end, instead of solely discussing the merging technique, we explore an \textbf{orthogonal} question: \textit{Can we scale up the size of model zoo to cover more models, and design an automatic merging technique to benefit from the inclusion?}

\vspace{-2mm}
\paragraph{Failure Case of Existing Approaches.} To begin with, we provide a motivating example to show the failure case of the existing approach. We consider the three models, Llama-2-Chat~\cite{Touvron2023Llama2O}, Vicuna~\cite{zheng2024judging} and CodeLlama~\cite{Rozire2023CodeLO}, all initialized with the same base model, Llama-2~\cite{Touvron2023Llama2O}. We merge Vicuna and CodeLlama with Llama-2-Chat, respectively, and report the evaluation results in Table~\ref{tab:mergability} in Appendix~\ref{sec:more_merging}. We evaluate $6$ representative merging techniques implemented in \textit{mergekit}~\cite{goddard2024arcee}, including linear interpolation~\cite{wortsman2022model},  SLERP~\cite{10.1145/325334.325242}, Model Stock~\cite{jang2024model}, Task Arithmetic~\cite{ilharco2023editing}, DARE~\cite{xu2024training}, and TIES~\cite{yadav2023tiesmerging}. By merging Llama-2-chat and Vicuna, the merged model achieves better performance compared to any single model, while merging Llama-2-chat and CodeLlama fails to outperform all single models and may even lead to a significant drop in performance, which is also mentioned by \citet{xu2024training}. The results indicate the potential severe performance drop when including un-mergeable new model in merging (e.g. CodeLlama). Even if it is obtained from the same pre-trained checkpoint. Such failure case motivates us to design the strategy to automatically select models for merging, and exclude the models that are unable to merge.

\vspace{-2mm}
In the following paragraphs, we explore several solutions tailored for large-scale model merging. These variations address different resource and speed requirements. 
The introduction of these methods is organized around answering the following key questions.

\vspace{-2mm}
\paragraph{\uline{Q1:} Does handcrafted rules apply to automated model selection and which one performs best? \\ A: Yes, by a greedy search approach.} In this section, we explore three potential heuristics for model selection and report the results in Figure~\ref{fig:merge_radar}(a). We include the performance of the ``best single model'' (the model participant before merging that achieves the best averaged performance). We additionally validate the performance of heuristic-based merging technique, which are detailed in Section~\ref{sec:merging_method}. As indicated by the results, the merging technique based on \textit{Heuristic-Coefficient} yields consistently superior performance when the model zoo is large. 
For \texttt{Which4}, \textit{Heuristic-Average} achieved better performance, while \textit{Heuristic-Coefficient} performed poorly. This is primarily because the domain-specific models in \texttt{Which4} exhibit similar performances and are indispensable. 
\vspace{-2mm}
\paragraph{\uline{Q2:} How to utilize Evolutionary Strategy for coefficient optimization in model merging? }\textcolor{white}{.}

\vspace{-7pt}
We divide the problem into the following sub-questions: 
(\textit{i}) Which merging method is most compatible with Evolutionary Strategy?
(\textit{ii}) Can finer-grained optimization lead to a better merged model?
(\textit{iii}) How to efficiently merge in a large model zoo?
For (\textit{i}), \textbf{A: simpler methods such as Linear and Task Arithmetic are more competitive.} We compared four methods: Linear, Task Arithmetic, DARE, and TIES. As shown in Figure~\ref{fig:merge_radar}(b), Linear merging consistently achieves great results. However, when the parameters to be optimized are small, Task Arithmetic performs slightly better than Linear. Under a fixed computational budget, due to the doubling of parameters to be optimized, DARE and TIES exhibit slightly lower performance compared to other methods. 
For (\textit{ii}), \textbf{A: Yes, but we need a larger computational budget.} We group adjacent $n$ decoder layers together, where they share the same coefficients. The group size $n\in [ 32, 8, 4, 1]$. 
When $n = 8$, better results were achieved compared to $n = 32$, as shown in Table~\ref{tab:evo-group-size}. However, as we further decreased the group size, the performance slightly declined. This could be attributed to our relatively small budget.
For (\textit{iii}), \textbf{A: Use Heuristic Strategy to roughly search for coefficients and then fine-tune the coefficients using Evolutionary Strategy.} As shown in Table~\ref{tab:efficient-merge}, the combination of the two strategies resulted in better results with fewer trials. For implementation details, please refer to Appendix~\ref{apdx:details_of_evo}.
\vspace{-5pt}

\subsection{Implementation Details for Mixture}\label{impl_mixture} 
\begin{wraptable}{r}{0.4\linewidth}
\centering
\vspace{-16pt}
\caption{\small Model mixture methods and their abbreviations used in our study. Methods applicable for models with distinct architectures are highlighted in \colorbox{gray!30}{gray}.}\label{tab:mixture-method}
\renewcommand\arraystretch{1}
\tabcolsep=0.1cm
\resizebox{\linewidth}{!}{
\begin{tabular}{r|cccc}
\toprule
\midrule
Abbreviation & Mix. Level & Router & Router Input & Hybrid \\ \midrule
F-L-T              &  FFN             &  Linear          &  Token   & \ding{55}     \\ 
Hybrid F-L-T             & FFN & Linear & Token & \ding{51}   \\
F-L-S               & FFN              & Linear           & Sample  & \ding{55}    \\ 
F-M-S               & FFN              & MLP           & Sample  & \ding{55}    \\ 
B-L-S               &  Block             &  Linear          & Sample & \ding{55}       \\ 
B-M-S               &  Block             &  MLP          & Sample & \ding{55}       \\ 
\rowcolor[gray]{0.9} 
M-L-S          &  Model             &  Linear          & Sample & \ding{55}       \\
\midrule
\bottomrule
\end{tabular}}
\vspace{-3mm}
\end{wraptable}
\vspace{-1mm}
\paragraph{Model Zoo and Router Initialization.} 

In Mixture Bench, we experiment with \texttt{Which2} and \texttt{Which4} model settings.
For router design, we mainly adopt a training-free linear layer router initialized from the prompt vector, as previous studies have demonstrated its effectiveness in the zero-shot MoE model~\cite{goddard2024arcee}. For specific prompt settings, we refer to the Beyonder model series~\footnote{\url{https://huggingface.co/mlabonne/Beyonder-4x7B-v2}}. For the routing algorithm, we use Top-$1$ routing for \texttt{Which2} and \textit{Block level mixture} and \textit{Model-level mixture} for \texttt{Which4}, and Top-$2$ for \texttt{Which4} \textit{FFN level mixture}.

\vspace{-2mm}
\paragraph{Post-mixture training.} For \textit{MLP router} that are randomly initialized, we fine-tune the model by language modeling on the GPT4All dataset ~\cite{peng2023instruction}, only updating the router. We use the GPT4All~\cite{peng2023instruction} dataset for post-mixture router training, which is under Apache 2.0 License. For all the router training experiments, we apply the batch size of $128$, a cosine learning rate scheduler, the learning rate of $5e-5$, and the epochs of $1$.


\vspace{-2mm}
\paragraph{Mixture Method Abbreviations.} To simplify the description, we use abbreviations to denote different mixture methods, as in Table~\ref{tab:mixture-method}.

\vspace{-2mm}
\subsection{Model Mixture Benchmark Results} \label{sec:mixture-results}

\vspace{-1mm}
In this section, we attempt to answer five main research questions about mixture variants: mixture level, router design, router input, and hybrid merging. We also explore the mixing of very different models that cannot be merged as the previous probe in our next \texttt{Model-GLUE} recipe that combines merging and blending for LLM scaling.  
\vspace{-1mm}
\paragraph{\uline{Q1:} At which level does the model mixture manifest its utmost effectiveness? } \textcolor{white}{.}

\begin{wraptable}{r}{0.7\linewidth}
\centering
\vspace{-18pt}
\caption{\small Comparison of different mixture levels. For each task in each model zoo, we highlight the performance best in each model zoo in \textbf{bold}.} \label{tab:q1-mixture-level}
\vspace{2pt}
\renewcommand\arraystretch{1}
\tabcolsep=0.1cm
\resizebox{1\linewidth}{!}{
\begin{tabular}{c|ccccccc}
\toprule
\midrule
Model & ARC  & WinoGrande  & MMLU & GSM8K & MBPP & HumanEval & Average \\ 
\midrule
\multicolumn{8}{c}{\texttt{Which2}} \\
\midrule
Best Single Model & $54.27\%$ & $71.51\%$ & $47.24\%$ & $21.30\%$ & $18.00\%$ & $13.06\%$ & $37.68\%$ \\ 
\midrule
F-L-S & $52.82\%$ & $70.80\%$ & $50.04\%$ & $\mathbf{23.12\%}$ & $19.00\%$ & $17.68\%$ & $38.91\%$  \\
B-L-S & $52.73\%$	& $70.01\%$	& $49.90\%$	& $19.94\%$	& $18.84\%$	& $15.85\%$	& $37.88\%$ \\
M-L-S         & $\mathbf{54.44\%}$       & $\mathbf{72.38\%}$    & $\mathbf{50.51\%}$    & $22.21\%$      & $\mathbf{20.00\%}$      & $\mathbf{20.73\%}$      & $\mathbf{40.04\%}$                \\
\midrule
\multicolumn{8}{c}{\texttt{Which4}} \\
\midrule
Best Single Model & $\mathbf{55.03\%}$ & $73.72\%$ & $\mathbf{48.33\%}$ & $24.26\%$ & $17.80\%$ & $13.41\%$ & $38.70\%$ \\  
\midrule
F-L-S  & $53.75\%$ & $73.88\%$ & $47.97\%$ & $34.87\%$ & $\mathbf{21.80\%}$ & $\mathbf{23.17\%}$ & $42.57\%$ \\ 
B-L-S  & $52.65\%$ & $\mathbf{74.66\%}$ & $47.05\%$ & $21.15\%$ & $20.40\%$ & $14.63\%$ & $38.42\%$ \\ 
M-L-S & $49.06\%$ & $72.14\%$ & $41.81\%$ & $\mathbf{60.05\%}$ & $17.60\%$ & $15.24\%$ & $\mathbf{42.65\%}$ \\ 
\midrule
\bottomrule
\end{tabular}}
\vspace{-15pt}
\end{wraptable}

\vspace{-4pt}
\textbf{A: Model level mixture is consistently better.} Our comparative analysis of the \{FFN, block, model\} level mixture, all employing the linear router and the sample routing strategy as presented in Table~\ref{tab:q1-mixture-level}, consistently demonstrates the superiority of the \textit{Model level mixture} under \texttt{Which2} and \texttt{Which4} setting. This could be attributed to the design that \textit{Model Level Mixture} route each sample to one expert model, thereby avoiding the conflicts between different expert models and maximizing the expertise of the most appropriate experts.
Since the experts are not derived from the same pre-training process, directly merging their inconsistent representation spaces will affect the performance of the mixture model, with more expert parameters leading to worse results. 
This is especially evident for \textit{Block-level Mixture}, as the routing is performed at each transformer layer and the representation is fed into different expert blocks in series, causing confusion when switching between different expert knowledge.

\vspace{-2mm}
\paragraph{\uline{Q2:} Does more complex router design brings better results?}\textcolor{white}{.}

\begin{wraptable}{r}{0.6\linewidth}
\centering
\vspace{-19pt}
\caption{\small Comparison between linear and MLP routers on \texttt{Which2} setting. We highlight better performance within each pair in \textbf{bold}.} \label{tab:q2-mixture-router}
\vspace{2pt}
\renewcommand\arraystretch{1}
\tabcolsep=0.1cm
\resizebox{1\linewidth}{!}{
\begin{tabular}{c|ccccccc}
\toprule
\midrule
Model & ARC  & WinoGrande  & MMLU & GSM8K & MBPP & HumanEval & Average \\ 
\midrule
F-L-T  &  $53.41\%$  &  $70.48\%$  &  $\mathbf{50.74}\%$       & $\mathbf{23.28\%}$ &  $\mathbf{20.80\%}$ & $16.46\%$  &  $\mathbf{39.20\%}$ \\
F-M-T         &  $\mathbf{53.58\%}$   & $\mathbf{72.06\%}$       &  $50.01\%$    & $21.92\%$      &  $17.40\%$     & $\mathbf{17.68\%}$     & $38.78\%$                \\
\midrule
B-L-S & $\mathbf{52.73\%}$	& $70.01\%$	& $\mathbf{49.90\%}$	& $\mathbf{19.94\%}$	& $\mathbf{18.84\%}$	& $\mathbf{15.85\%}$	& $\mathbf{37.88\%}$ \\
B-M-S   & $51.53\%$ & $\mathbf{70.56\%}$ & $49.41\%$& $\mathbf{19.94\%}$ & $16.60\%$& $14.02\%$ & $37.01\%$ \\
\midrule
\bottomrule
\end{tabular}}
\vspace{-12pt}
\end{wraptable}

\vspace{-4pt}
\textbf{A: Not necessary, as the linear router outperforms the MLP router.} From Table ~\ref{tab:q2-mixture-router}, the performances of the \textit{linear router} without additional training slightly surpass \textit{MLP router} models, \textit{i.e.}, F-L-T over F-M-T, B-L-S over B-M-S. 
Specifically, \textit{linear router} models are better at math and coding datasets, validating prompt vector is effective in assorting samples from different domains, which is otherwise too implicit to learn via direct language modeling. 
\vspace{-6mm}
\paragraph{\uline{Q3:} Does model mixture directly works  on unmergeable models?}\textcolor{white}{.} 

\begin{wraptable}{r}{0.7\linewidth}
\centering
\vspace{-18pt}
\caption{\small Comparison of the mixture of a unmergeable model pair (Llama-2-7b-chat and CrystalChat). We highlight the better performance in \textbf{bold}.} \label{tab:q5-diverse-architecture}
\renewcommand\arraystretch{1}
\tabcolsep=0.1cm
\resizebox{1\linewidth}{!}{
\begin{tabular}{c|ccccccc}
\toprule
\midrule
Model & ARC  & WinoGrande  & MMLU & GSM8K & MBPP & HumanEval & Average \\ 
\midrule
Best Single Model & $\mathbf{52.05\%}$ & $69.46\%$ & $\mathbf{50.77\%}$ & $27.22\%$ & $\mathbf{39.60\%}$ & $\mathbf{35.98\%}$ & $\mathbf{45.85\%}$ \\ 
\midrule
M-L-S  &  $50.68\%$  &  $\mathbf{69.77\%}$  &  $50.08\%$       & $\mathbf{27.82\%}$  &  $33.80\%$ & $30.48\%$  & $43.77\%$   \\
\midrule
\bottomrule
\end{tabular}}
\vspace{-10pt}
\end{wraptable}
\vspace{-1mm}
\textbf{A: No.} We directly apply the setting of \texttt{Which2} \textit{Model level mixture} to Llama-2-7b-chat and CrystalChat, an unmergeable model pair with different architectures and initialization. As shown in Table~\ref{tab:q5-diverse-architecture}, the performance is slightly behind the best single model. This may be due to simple prompts and direct mixture, as it fails to coordinate the divergence between drastically different models. We evaluate more complex prompts for the same model pair and the mixture model outperforms, see Table~\ref{tab:q5-diverse-architecture-2} for more information. 


\vspace{-2mm}
\paragraph{\uline{Q4:} Which router input is better, token-level or sample-level? }\textcolor{white}{.} 

\begin{wraptable}{r}{0.7\linewidth}
\centering
\vspace{-19pt}
\caption{\small Comparison of different router input designs. \texttt{Which4} includes one group with chatting models \texttt{(Chat)} and another with different domain models \texttt{(Domain)} . We highlight the best performing mixture methods in \textbf{bold}. } \label{tab:q3-router-input}
\vspace{3pt}
\renewcommand\arraystretch{1}
\tabcolsep=0.1cm
\resizebox{1\linewidth}{!}{
\begin{tabular}{c|ccccccc}
\toprule
\midrule
Model & ARC  & WinoGrande  & MMLU & GSM8K & MBPP & HumanEval & Average \\ 
\midrule
\multicolumn{8}{c}{\texttt{Which2}} \\
\midrule
Best Single Model & $\mathbf{54.27\%}$ & $\mathbf{71.51}\%$ & $47.24\%$ & $21.30\%$ & $18.00\%$ & $13.06\%$ & $37.68\%$ \\ 
\midrule
F-L-T  &  $53.41\%$  &  $70.48\%$  &  $\mathbf{50.74}\%$       & $\mathbf{23.28\%}$ &  $\mathbf{20.80\%}$ & $16.46\%$  &  $\mathbf{39.20\%}$ \\
F-L-S & $52.82\%$ & $70.80\%$ & $50.04\%$ & $23.12\%$ & $19.00\%$ & $\mathbf{17.68\%}$ & $38.91\%$  \\
\midrule
\multicolumn{8}{c}{\texttt{Which4}} \\
\midrule
Best Single Model & $55.03\%$ & $73.72\%$ & $\mathbf{48.33\%}$ & $24.26\%$ & $17.80\%$ & $13.41\%$ & $38.70\%$ \\  
\midrule
Chat F-L-T & $\mathbf{55.63}\%$ & $\mathbf{72.77}\%$ & $\mathbf{50.28}\%$ & $\mathbf{23.88}\%$ & $20.00\%$ & $\mathbf{22.56}\%$ & $\mathbf{40.85}\%$ \\ 
Chat F-L-S & $53.75\%$ & $70.96\%$ & $49.78\%$ & $20.32\%$ & $\mathbf{20.40}\%$ & $20.12\%$ & $39.22\%$ \\ \midrule
Domain F-L-T & $\mathbf{55.72\%}$ & $\mathbf{74.11\%}$  & $48.32\%$  & $30.17\%$ & $\mathbf{22.00\%}$ & $20.12\%$ & $41.74\%$ \\ 
Domain F-L-S  & $53.75\%$ & $73.88\%$ & $47.97\%$ & $\mathbf{34.87\%}$ & $21.80\%$ & $\mathbf{23.17\%}$ & $\mathbf{42.57\%}$ \\ 
\midrule
\bottomrule
\end{tabular}}
\vspace{-12pt}
\end{wraptable}

\vspace{-2mm}
\textbf{A: Not quite different. Token input suits a mixture of the same domain models.} Table~\ref{tab:q3-router-input} shows the performance token-based and sample-based routing are pretty close. In particular, for \texttt{Which2} and \texttt{Which4 (Chat)} where models are all trained for general chatting purposes, token routing outperforms, whereas sample routing is better for default \texttt{Which4 (Domain)} with differently specialized models. This may result from divergence of model knowledge and representation spaces will cause conflicts in fine-grained token routing.

\vspace{-2mm}
\paragraph{\uline{Q5:} Is it feasible for hybrid mixtures to provide enhancements?} \textcolor{white}{.} 

\begin{wraptable}{r}{0.7\linewidth}
\centering
\vspace{-20pt}
\caption{\small Comparison between F-L-T methods with and without hybrid mixture technique. We highlight the best performing mixture methods in \textbf{bold}. } \label{tab:q4-hybrid}
\renewcommand\arraystretch{1}
\tabcolsep=0.1cm
\resizebox{1\linewidth}{!}{
\begin{tabular}{c|ccccccc}
\toprule
\midrule
Model & ARC  & WinoGrande  & MMLU & GSM8K & MBPP & HumanEval & Average \\ 
\midrule
\multicolumn{8}{c}{\texttt{Which2}} \\
\midrule
Best Single Model & $54.27\%$ & $\mathbf{71.51\%}$ & $47.24\%$ & $21.30\%$ & $18.00\%$ & $13.06\%$ & $37.68\%$ \\ 
\midrule
F-L-T  &  $53.41\%$  &  $70.48\%$  &  $\mathbf{50.74}\%$       & $23.28\%$ &  $20.80\%$ & $16.46\%$  &  $39.20\%$ \\
Hybrid F-L-T  & $\mathbf{54.44\%}$       & $71.19\%$     & $50.45\%$    & $\mathbf{23.96\%}$      & $\mathbf{21.80\%}$      & $\mathbf{18.29\%}$      & $\mathbf{40.02\%}$ \\
\midrule
\multicolumn{8}{c}{\texttt{Which4}} \\
\midrule
Best Single Model & $55.03\%$ & $73.72\%$ & $\mathbf{48.33\%}$ & $24.26\%$ & $17.80\%$ & $13.41\%$ & $38.70\%$ \\  
\midrule
F-L-T & $\mathbf{55.72\%}$ & $\mathbf{74.11\%}$  & $48.32\%$  & $30.17\%$ & $22.00\%$ & $20.12\%$ & $41.74\%$ \\  
Hybrid F-L-T & $54.86\%$ & $73.80\%$ & $48.23\%$ & $\mathbf{37.53\%}$ & $\mathbf{24.30\%}$ & $\mathbf{23.17\%}$ & $\mathbf{43.65\%}$\\
\midrule
\bottomrule
\end{tabular}}
\vspace{-12pt}
\end{wraptable}

\vspace{-4pt}
\textbf{A: Yes.} Our experiments on F-L-T with \textit{v.s.} without the hybrid mixture, as detailed in Table~\ref{tab:q4-hybrid}, demonstrate that the hybrid mixture significantly improves performance on average and simultaneously reduces the memory overhead during inference. This improvement may be attributed to the higher sensitivity of the initial transformer blocks. Avoiding using MoE for these blocks can yield performance gains, as suggested by a few previous works as well~\cite{dai2024deepseekmoe, rajbhandari2022deepspeedmoe}. Surprisingly, our results show that the hybrid F-L-T model consistently outperforms the standard F-L-T on \textit{math} and \textit{code} tasks. Our further analysis indicates that this improvement might be because of the conversational nature of the content in GSM8K, MBPP, and HumanEval datasets, which appears to challenge the routing mechanisms within the initial transformer blocks, leading to ineffective expert specialization.

\section{Superior Recipes to Aggregate LLM Knowledge}

\subsection{Model Merging \textit{v.s.} Mixture}\label{sec:merging_vs_misture}
\paragraph{\uline{Q1:} For a mergeable model zoo, how should we choose between merging and mixture?}
For limited computational resources and similar models, merging is always a simple and effective method. For the domain-specific models, mixture can bring greater improvements.

\begin{wraptable}{r}{0.65\linewidth}
\centering
\vspace{-18pt}
\caption{\small Comparison between the best merging approach \textit{v.s.} the best mixture approach on \texttt{Which4~(Domain) and \texttt{Which4~(Chat)}}.} \label{tab:best merging vs. best mixture}
\renewcommand\arraystretch{1}
\tabcolsep=0.1cm
\resizebox{1\linewidth}{!}{
\begin{tabular}{c|cccccccc}
\midrule
\midrule
Model & ARC  & WinoGrande  & MMLU & GSM8K & MBPP & HumanEval & Average \\ \midrule
Best Single Model       & $55.03\% $         & $73.72\%$          & $48.18\%$          & $24.03\%$          & $17.80\%$          & $13.41\%$          & $38.70\%$          \\
\midrule
\multicolumn{8}{c}{\texttt{Which4~(Domain)}} \\
\midrule
Merging & $54.01\%$          & $73.64\%$ & $47.39\% $         & $43.75\%$          & $\mathbf{22.40}\%$          & $\mathbf{21.95}\%$ & $43.86\%$  \\ 
Mixture &  $\mathbf{54.86}\%$	&  $\mathbf{74.11}\%$	&  $\mathbf{48.23}\%$	&  $\mathbf{49.81}\%$	&  $18.40\%$	&  $18.29\%$	&  $\mathbf{43.95}\%$ \\
\midrule
\multicolumn{8}{c}{\texttt{Which4~(Chat)}} \\
\midrule
Merging & $\mathbf{56.23}\%$        & $\mathbf{73.72}\%$           & $\mathbf{50.51}\%$        & $\mathbf{25.85}\%$        & $\mathbf{21.00}\%$        & $\mathbf{21.95}\%$          & $\mathbf{41.54}\%$       \\
Mixture & $53.75\%$ &	$70.96\%$ & $49.80\%$	& $19.94\%$	& $19.80\%$	& $20.73\%$	& $39.16\%$ \\
\midrule
\bottomrule
\end{tabular}}
\vspace{-10pt}
\end{wraptable}
\vspace{-2mm}
Detailed results are presented in Table~\ref{tab:best merging vs. best mixture}. \uline{For \texttt{Which4}~(Domain)}, due to the appropriately designed linear routers, model mixture can fully leverage various domain-specific models, thus slightly outperforming merging. 
\uline{For \texttt{Which4~(Chat)}}, we adopt the optimal settings from \texttt{Which4~(Domain)} and only change the model zoo.
Since individual models do not exhibit superior capabilities in a single domain, it is challenging to design suitable routers at a low cost. 
Therefore, mixture performed significantly worse compared to merging.
Furthermore, although combining the homogeneous models in \texttt{Which4}~(Chat) brings some improvement, we can see that \texttt{Which4~(Domain)} overall outperforms \texttt{Which4~(Chat)}. Therefore, increasing the diversity among the models will make a greater contribution to the combined model.

\subsection{\texttt{Model-GLUE}: selective merging then model mixture for better LLM scaling}\label{sec:model_glue}
\paragraph{\uline{Q2:} How to combine models with greater differences in an extensive and varied model zoo?} \textcolor{white}{...} 

In \texttt{Which16}, a larger and more diverse model zoo , some models cannot be merged due to structural differences and models that would degrade in performance when merged with other models. 
Therefore, we first cluster the models based on cosine similarity. 
Within each mergeable family, we perform either merging or mixture. 
We initially employ heuristic strategies of merging and report the best results (\textit{i.e.}, \texttt{Full Merging}) in Table~\ref{tab:model_glue}. 
The Llama-2 family (\textit{i.e.}, \texttt{Which12}) consists of up to $12$ models, so directly combining them through the mixture is inefficient. 
Thus, we only consider models selected by merging and report the results of \texttt{F-L-T Mixture}.
From Table~\ref{tab:model_glue}, we can observe that \texttt{Full Merging} outperforms \texttt{F-L-T Mixture}. 

\begin{wraptable}{r}{0.65\linewidth}
\centering
\vspace{-19pt}
\caption{\small Comparison between the best single model, Full Merging, Full Mixture and our \texttt{Model-GLUE}.} \label{tab:model_glue}
\vspace{2pt}
\renewcommand\arraystretch{1}
\tabcolsep=0.1cm
\resizebox{1\linewidth}{!}{
\begin{tabular}{c|cccccccc}
\toprule
\midrule
Model & ARC  & WinoGrande  & MMLU & GSM8K & MBPP & HumanEval & Average \\ \midrule
Best Single Model & $46.76\%$ & $64.33\%$ & $46.33\%$ & $\mathbf{62.40\%}$ & $42.00\%$ & $31.10\%$ & $48.82\%$ \\ \midrule
Full Merging & $\mathbf{55.12\%}$ & $\mathbf{73.64}\%$ & $50.13\%$ & $39.35\%$ & $21.80\%$ & $21.34\%$ & $43.56\%$  \\ 
F-L-T Mixture  & $54.69\%$                     & $73.32\%$                     & $48.74\%$ & $35.18\%$                     & $22.60\%$                     & $21.34\%$                     & $42.65\%$   \\ \midrule
\texttt{Model-GLUE} & $51.62\%$ & $70.56\%$ &	$\mathbf{51.85\%}$ &	$53.53\%$ &	$\mathbf{47.20\%}$ &	$\mathbf{51.83\%}$ &	$\mathbf{54.43\%}$ \\
\midrule
\bottomrule
\end{tabular}}
\vspace{-15pt}
\end{wraptable}

\vspace{-2mm}
Therefore, we selected \texttt{Full Merging} as the representative model for the Llama-2 family and combined it with other models that could not be merged by model mixture. On average, the \texttt{Model-GLUE} demonstrates a $5.61\%$ improvement over the \texttt{Best Single Model}. More details are presented in Appendix~\ref{apdx:modelglue}.

\section{Discussion with Other LLM Aggregation Techniques}\label{sec6}
Thus far, we mainly focus on two LLM aggregation techniques: model merging and mixture. In this section, we discuss other potential techniques that could help scaling existing LLMs.

\vspace{-2mm}
\paragraph{Model Stacking.}
Research has demonstrated that stacking a model itself can accelerate training convergence as opposed to training a model of double the size from scratch \cite{gong2019efficient, gu2020transformer, wang2023learning, wu2024llama, kim2023solar}.
This concept can be extended naturally to stack multiple models as one larger model.
Our experimental results indicate that model stacking with lightweight fine-tuning can yield superior performance compared to various merging and mixture models.
For instance, stacking 7B Llama-2-chat and Vicuna can achieve $\ge55\%$ on the MMLU benchmark.
When compared to model mixture, model stacking offers less flexibility in terms of model choices.
Although the resulting architecture is more standardized than MoE, increasing the model depth through stacking also results in higher latency than mixture models where subnetworks infer in parallel. Additionally, model stacking does not simplify the design space, such as determining whether, which, and how many layers should be dropped when stacking two heterogeneous models.
We conducted a preliminary investigation employing model stacking techniques to address two primary research questions:
\uline{($1$)} Can model stacking effectively combine the capabilities of two distinct models and surpass the performance of self-stacking a single model?
\uline{($2$)} What is the impact of layer dropping on stacking performance? 

Specifically, we examine the relationship between the number of dropped layers ($K$) and the resulting downstream task accuracy.
To this end, we selected 7B Llama-2-Chat and Vicuna as the base models and fine-tuned the stacked models for 10 billion tokens. The obtained results are presented in Table \ref{table:stacking_perf}. In the initial two rows, we report the performance of the two base models, revealing that Llama and Vicuna exhibit advantages on different datasets. In the subsequent two rows, we observe that stacking dissimilar models generally outperforms self-stacked models, and the weaknesses of one model can be compensated for by another stronger one. Moving forward, we explored the effects of varying the number of dropped layers. Our findings indicate that even when dumping half of each model ($K=16$), the stacked 7B models can still significantly enhance performance across tasks.

\begin{table}[h!]
\centering
\caption{Comparison of different model stacking configurations.}
\vspace{-2mm}
\label{table:stacking_perf}
\resizebox{0.8\linewidth}{!}{
\begin{tabular}{l|ccccc}
\toprule
\midrule
Model & ARC & WinoGrande& MMLU  & Hellaswag & TruthfulQA  \\ \hline
Llama-2-chat        & $54.10\%$                      & $71.27\%$   & $47.28\%$          & $78.71\%$             & $45.32\%$               \\
Vicuna       & $53.75\%$                  & $70.56\%$       &     $49.78\%$       & $77.19\%$          & $50.36\%$                \\
\hline
Llama / Llama ($K=8$)            & $53.92\%$                    & $69.14\%$        & $52.76\%$       & $73.74\%$             & $46.36\%$          \\
Llama / Vicuna ($K=8$)         & $56.14\%$                      & $70.80\%$    & $55.20\%$     & $73.67\%$             & $46.84\%$           \\
\hline
Llama / Vicuna ($K=12$)        & $55.42\%$              & $69.45\%$         & $53.55\%$        & $73.62\%$             & $45.59\%$           \\
Llama / Vicuna ($K=16$)        & $54.35\%$                 & $69.69\%$        & $52.52\%$          & $73.75\%$             & $45.92\%$        \\
Llama / Vicuna ($K=20$)        & $39.59\%$                    & $61.33\%$  & $44.93\%$         & $62.10\%$             & $42.90\%$           \\
Llama / Vicuna ($K=24$)        & $28.15\%$                & $52.88\%$        & $25.51\%$         & $43.07\%$             & $39.10\%$            \\
\midrule
\bottomrule
\end{tabular}}
\end{table}

\vspace{-2mm}
\paragraph{Model Communication.}
Model communication \cite{wu2023autogen, li2024camel, liang2023encouraging} is a framework that enables the development of LLM applications through the use of multiple conversable agents that collaborate to complete tasks.
This approach allows developers to design complex LLM application workflows as multi-agent conversations, where agents with various roles and capabilities, driven by LLMs, tools, or human inputs, interact with each other.
Unlike model merging, mixture, and stacking techniques, LLM communication is orthogonal to the primary focus of this paper because it does not modify the model weights; instead, it leverages the in-context learning and conversational capabilities of LLMs to coordinate agents.
An empirical comparison with this class of methods is beyond the scope of this study and will be explored in future research.

\vspace{-1mm}
\section{Limitations}
\vspace{-2mm}
For LLM scaling studies, while empirical evidence suggests that increasing model size, data volume, and computational complexity leads to better performance, there is little theoretical clarity on the exact mechanisms behind these improvements. Second, although scaling laws suggest that performance continues to improve as models get larger, recent evidence indicates that scaling may lead to diminishing returns beyond a certain point. In addition, our work focuses on benchmarking results, while the reasons why model merging improves performance could be further enhanced by post hoc analysis, such as examining parameter distribution and similarity during model operations.

\vspace{-1mm}
\section{Conclusion}\label{conclusion}
\vspace{-2mm}
In this paper, we explore the scaling LLM based on a model zoo of pre-trained LLMs within the real world. We first benchmark state-of-the-art LLM merging, mixture, and model stacking. Based on previous findings, we then propose a novel LLM scaling framework, \texttt{Model-GLUE}. Specifically, we scale up the model zoo closely examine the existing model merging techniques, and conclude the selective merging techniques based on heuristics and learnable algorithms. Further, we investigate variants of Mixture-of-Experts for combining LLMs and suggest it can serve as an alternative to merging failure cases. Finally, we integrate selective merging strategies with model mixture techniques, presenting this as a comprehensive solution for scaling a diverse array of LLM collections. Future works will include model stacking and communication to our \texttt{Model-GLUE} framework.

\bibliographystyle{plainnat}
\newpage
\bibliography{modelglue}

\begin{thebibliography}{68}
\providecommand{\natexlab}[1]{#1}
\providecommand{\url}[1]{\texttt{#1}}
\expandafter\ifx\csname urlstyle\endcsname\relax
  \providecommand{\doi}[1]{doi: #1}\else
  \providecommand{\doi}{doi: \begingroup \urlstyle{rm}\Url}\fi

\bibitem[Agarwal et~al.(2024)Agarwal, Awasthi, Kale, and Zhao]{agarwal2024stacking}
Naman Agarwal, Pranjal Awasthi, Satyen Kale, and Eric Zhao.
\newblock Stacking as accelerated gradient descent.
\newblock \emph{arXiv preprint arXiv:2403.04978}, 2024.

\bibitem[Ainsworth et~al.(2022)Ainsworth, Hayase, and Srinivasa]{ainsworth2022git}
Samuel~K Ainsworth, Jonathan Hayase, and Siddhartha Srinivasa.
\newblock Git re-basin: Merging models modulo permutation symmetries.
\newblock \emph{arXiv preprint arXiv:2209.04836}, 2022.

\bibitem[Akiba et~al.(2019)Akiba, Sano, Yanase, Ohta, and Koyama]{akiba2019optuna}
Takuya Akiba, Shotaro Sano, Toshihiko Yanase, Takeru Ohta, and Masanori Koyama.
\newblock Optuna: A next-generation hyperparameter optimization framework.
\newblock In \emph{Proceedings of the 25th ACM SIGKDD international conference on knowledge discovery \& data mining}, pages 2623--2631, 2019.

\bibitem[Akiba et~al.(2024)Akiba, Shing, Tang, Sun, and Ha]{akiba2024evolutionary}
Takuya Akiba, Makoto Shing, Yujin Tang, Qi~Sun, and David Ha.
\newblock Evolutionary optimization of model merging recipes, 2024.

\bibitem[Anand et~al.(2023)Anand, Nussbaum, Duderstadt, Schmidt, Treat, and Mulyar]{peng2023instruction}
Yuvanesh Anand, Zach Nussbaum, Brandon Duderstadt, Benjamin~M. Schmidt, Adam Treat, and Andriy Mulyar.
\newblock Gpt4all-j: An apache-2 licensed assistant-style chatbot, 2023.

\bibitem[Austin et~al.(2021)Austin, Odena, Nye, Bosma, Michalewski, Dohan, Jiang, Cai, Terry, Le, and Sutton]{Austin2021ProgramSW}
Jacob Austin, Augustus Odena, Maxwell Nye, Maarten Bosma, Henryk Michalewski, David Dohan, Ellen Jiang, Carrie~J. Cai, Michael Terry, Quoc~V. Le, and Charles Sutton.
\newblock Program synthesis with large language models.
\newblock \emph{ArXiv}, abs/2108.07732, 2021.

\bibitem[Azerbayev et~al.(2023)Azerbayev, Schoelkopf, Paster, Santos, McAleer, Jiang, Deng, Biderman, and Welleck]{azerbayev2023llemma}
Zhangir Azerbayev, Hailey Schoelkopf, Keiran Paster, Marco~Dos Santos, Stephen McAleer, Albert~Q Jiang, Jia Deng, Stella Biderman, and Sean Welleck.
\newblock Llemma: An open language model for mathematics.
\newblock \emph{arXiv preprint arXiv:2310.10631}, 2023.

\bibitem[Bekbayev et~al.(2023)Bekbayev, Chun, Dulat, and Yamazaki]{bekbayev2023poison}
Aibek Bekbayev, Sungbae Chun, Yerzat Dulat, and James Yamazaki.
\newblock The poison of alignment, 2023.

\bibitem[Chen et~al.(2021)Chen, Tworek, Jun, Yuan, Ponde, Kaplan, Edwards, Burda, Joseph, Brockman, Ray, Puri, Krueger, Petrov, Khlaaf, Sastry, Mishkin, Chan, Gray, Ryder, Pavlov, Power, Kaiser, Bavarian, Winter, Tillet, Such, Cummings, Plappert, Chantzis, Barnes, Herbert-Voss, Guss, Nichol, Babuschkin, Balaji, Jain, Carr, Leike, Achiam, Misra, Morikawa, Radford, Knight, Brundage, Murati, Mayer, Welinder, McGrew, Amodei, McCandlish, Sutskever, and Zaremba]{Chen2021EvaluatingLL}
Mark Chen, Jerry Tworek, Heewoo Jun, Qiming Yuan, Henrique Ponde, Jared Kaplan, Harrison Edwards, Yura Burda, Nicholas Joseph, Greg Brockman, Alex Ray, Raul Puri, Gretchen Krueger, Michael Petrov, Heidy Khlaaf, Girish Sastry, Pamela Mishkin, Brooke Chan, Scott Gray, Nick Ryder, Mikhail Pavlov, Alethea Power, Lukasz Kaiser, Mohammad Bavarian, Clemens Winter, Philippe Tillet, Felipe~Petroski Such, David~W. Cummings, Matthias Plappert, Fotios Chantzis, Elizabeth Barnes, Ariel Herbert-Voss, William~H. Guss, Alex Nichol, Igor Babuschkin, Suchir Balaji, Shantanu Jain, Andrew Carr, Jan Leike, Joshua Achiam, Vedant Misra, Evan Morikawa, Alec Radford, Matthew~M. Knight, Miles Brundage, Mira Murati, Katie Mayer, Peter Welinder, Bob McGrew, Dario Amodei, Sam McCandlish, Ilya Sutskever, and Wojciech Zaremba.
\newblock Evaluating large language models trained on code.
\newblock \emph{ArXiv}, abs/2107.03374, 2021.

\bibitem[Clark et~al.(2018)Clark, Cowhey, Etzioni, Khot, Sabharwal, Schoenick, and Tafjord]{Clark2018ThinkYH}
Peter Clark, Isaac Cowhey, Oren Etzioni, Tushar Khot, Ashish Sabharwal, Carissa Schoenick, and Oyvind Tafjord.
\newblock Think you have solved question answering? try arc, the ai2 reasoning challenge.
\newblock \emph{ArXiv}, abs/1803.05457, 2018.

\bibitem[Cobbe et~al.(2021)Cobbe, Kosaraju, Bavarian, Chen, Jun, Kaiser, Plappert, Tworek, Hilton, Nakano, Hesse, and Schulman]{Cobbe2021TrainingVT}
Karl Cobbe, Vineet Kosaraju, Mohammad Bavarian, Mark Chen, Heewoo Jun, Lukasz Kaiser, Matthias Plappert, Jerry Tworek, Jacob Hilton, Reiichiro Nakano, Christopher Hesse, and John Schulman.
\newblock Training verifiers to solve math word problems.
\newblock \emph{ArXiv}, abs/2110.14168, 2021.

\bibitem[Dai et~al.(2024)Dai, Deng, Zhao, Xu, Gao, Chen, Li, Zeng, Yu, Wu, Xie, Li, Huang, Luo, Ruan, Sui, and Liang]{dai2024deepseekmoe}
Damai Dai, Chengqi Deng, Chenggang Zhao, R.~X. Xu, Huazuo Gao, Deli Chen, Jiashi Li, Wangding Zeng, Xingkai Yu, Y.~Wu, Zhenda Xie, Y.~K. Li, Panpan Huang, Fuli Luo, Chong Ruan, Zhifang Sui, and Wenfeng Liang.
\newblock Deepseekmoe: Towards ultimate expert specialization in mixture-of-experts language models, 2024.

\bibitem[Ding et~al.(2024)Ding, Chen, Cui, Lv, Zhao, Xie, Zhou, Liu, and Sun]{ding2024mastering}
Ning Ding, Yulin Chen, Ganqu Cui, Xingtai Lv, Weilin Zhao, Ruobing Xie, Bowen Zhou, Zhiyuan Liu, and Maosong Sun.
\newblock Mastering text, code and math simultaneously via fusing highly specialized language models, 2024.

\bibitem[Dodge et~al.(2022)Dodge, Prewitt, des Combes, Odmark, Schwartz, Strubell, Luccioni, Smith, DeCario, and Buchanan]{Dodge2022MeasuringTC}
Jesse Dodge, Taylor Prewitt, R{\'e}mi~Tachet des Combes, Erika Odmark, Roy Schwartz, Emma Strubell, Alexandra~Sasha Luccioni, Noah~A. Smith, Nicole DeCario, and Will Buchanan.
\newblock Measuring the carbon intensity of ai in cloud instances.
\newblock \emph{Proceedings of the 2022 ACM Conference on Fairness, Accountability, and Transparency}, 2022.

\bibitem[Fedus et~al.(2022)Fedus, Zoph, and Shazeer]{switch}
William Fedus, Barret Zoph, and Noam Shazeer.
\newblock Switch transformers: Scaling to trillion parameter models with simple and efficient sparsity.
\newblock \emph{J. Mach. Learn. Res.}, 23:\penalty0 120:1--120:39, 2022.

\bibitem[Goddard et~al.(2024)Goddard, Siriwardhana, Ehghaghi, Meyers, Karpukhin, Benedict, McQuade, and Solawetz]{goddard2024arcee}
Charles Goddard, Shamane Siriwardhana, Malikeh Ehghaghi, Luke Meyers, Vlad Karpukhin, Brian Benedict, Mark McQuade, and Jacob Solawetz.
\newblock Arcee's mergekit: A toolkit for merging large language models.
\newblock \emph{arXiv preprint arXiv:2403.13257}, 2024.

\bibitem[Gong et~al.(2019)Gong, He, Li, Qin, Wang, and Liu]{gong2019efficient}
Linyuan Gong, Di~He, Zhuohan Li, Tao Qin, Liwei Wang, and Tieyan Liu.
\newblock Efficient training of bert by progressively stacking.
\newblock In \emph{International conference on machine learning}, pages 2337--2346. PMLR, 2019.

\bibitem[Gu et~al.(2020)Gu, Liu, Yu, Li, Chen, and Han]{gu2020transformer}
Xiaotao Gu, Liyuan Liu, Hongkun Yu, Jing Li, Chen Chen, and Jiawei Han.
\newblock On the transformer growth for progressive bert training.
\newblock \emph{arXiv preprint arXiv:2010.12562}, 2020.

\bibitem[Hansen(2006)]{hansen2006cma}
Nikolaus Hansen.
\newblock The cma evolution strategy: a comparing review.
\newblock \emph{Towards a new evolutionary computation: Advances in the estimation of distribution algorithms}, pages 75--102, 2006.

\bibitem[Hendrycks et~al.(2020)Hendrycks, Burns, Basart, Zou, Mazeika, Song, and Steinhardt]{Hendrycks2020MeasuringMM}
Dan Hendrycks, Collin Burns, Steven Basart, Andy Zou, Mantas Mazeika, Dawn~Xiaodong Song, and Jacob Steinhardt.
\newblock Measuring massive multitask language understanding.
\newblock \emph{ArXiv}, abs/2009.03300, 2020.

\bibitem[Hu et~al.(2022)Hu, Shen, Wallis, Allen-Zhu, Li, Wang, Wang, and Chen]{hu2022lora}
Edward~J Hu, Yelong Shen, Phillip Wallis, Zeyuan Allen-Zhu, Yuanzhi Li, Shean Wang, Lu~Wang, and Weizhu Chen.
\newblock Lo{RA}: Low-rank adaptation of large language models.
\newblock In \emph{International Conference on Learning Representations}, 2022.
\newblock URL \url{https://openreview.net/forum?id=nZeVKeeFYf9}.

\bibitem[Ilharco et~al.(2023)Ilharco, Ribeiro, Wortsman, Gururangan, Schmidt, Hajishirzi, and Farhadi]{ilharco2023editing}
Gabriel Ilharco, Marco~Tulio Ribeiro, Mitchell Wortsman, Suchin Gururangan, Ludwig Schmidt, Hannaneh Hajishirzi, and Ali Farhadi.
\newblock Editing models with task arithmetic, 2023.

\bibitem[Imfeld et~al.(2023)Imfeld, Graldi, Giordano, Hofmann, Anagnostidis, and Singh]{imfeld2023transformer}
Moritz Imfeld, Jacopo Graldi, Marco Giordano, Thomas Hofmann, Sotiris Anagnostidis, and Sidak~Pal Singh.
\newblock Transformer fusion with optimal transport.
\newblock \emph{arXiv preprint arXiv:2310.05719}, 2023.

\bibitem[Jang et~al.(2024)Jang, Yun, and Han]{jang2024model}
Dong-Hwan Jang, Sangdoo Yun, and Dongyoon Han.
\newblock Model stock: All we need is just a few fine-tuned models, 2024.

\bibitem[Jiang et~al.(2024)Jiang, Sablayrolles, Roux, Mensch, Savary, Bamford, Chaplot, de~las Casas, Hanna, Bressand, Lengyel, Bour, Lample, Lavaud, Saulnier, Lachaux, Stock, Subramanian, Yang, Antoniak, Scao, Gervet, Lavril, Wang, Lacroix, and Sayed]{jiang2024mixtral}
Albert~Q. Jiang, Alexandre Sablayrolles, Antoine Roux, Arthur Mensch, Blanche Savary, Chris Bamford, Devendra~Singh Chaplot, Diego de~las Casas, Emma~Bou Hanna, Florian Bressand, Gianna Lengyel, Guillaume Bour, Guillaume Lample, Lélio~Renard Lavaud, Lucile Saulnier, Marie-Anne Lachaux, Pierre Stock, Sandeep Subramanian, Sophia Yang, Szymon Antoniak, Teven~Le Scao, Théophile Gervet, Thibaut Lavril, Thomas Wang, Timothée Lacroix, and William~El Sayed.
\newblock Mixtral of experts, 2024.

\bibitem[Jin et~al.(2023)Jin, Ren, Preotiuc-Pietro, and Cheng]{jin2023dataless}
Xisen Jin, Xiang Ren, Daniel Preotiuc-Pietro, and Pengxiang Cheng.
\newblock Dataless knowledge fusion by merging weights of language models, 2023.

\bibitem[Kaplan et~al.(2020)Kaplan, McCandlish, Henighan, Brown, Chess, Child, Gray, Radford, Wu, and Amodei]{Kaplan2020ScalingLF}
Jared Kaplan, Sam McCandlish, Tom Henighan, Tom~B. Brown, Benjamin Chess, Rewon Child, Scott Gray, Alec Radford, Jeff Wu, and Dario Amodei.
\newblock Scaling laws for neural language models.
\newblock \emph{ArXiv}, abs/2001.08361, 2020.

\bibitem[Kim et~al.(2023)Kim, Park, Kim, Lee, Song, Kim, Kim, Kim, Lee, Kim, et~al.]{kim2023solar}
Dahyun Kim, Chanjun Park, Sanghoon Kim, Wonsung Lee, Wonho Song, Yunsu Kim, Hyeonwoo Kim, Yungi Kim, Hyeonju Lee, Jihoo Kim, et~al.
\newblock Solar 10.7 b: Scaling large language models with simple yet effective depth up-scaling.
\newblock \emph{arXiv preprint arXiv:2312.15166}, 2023.

\bibitem[Lee et~al.(2023)Lee, Hunter, and Ruiz]{platypus2023}
Ariel~N. Lee, Cole~J. Hunter, and Nataniel Ruiz.
\newblock Platypus: Quick, cheap, and powerful refinement of llms.
\newblock 2023.

\bibitem[Lepikhin et~al.(2020)Lepikhin, Lee, Xu, Chen, Firat, Huang, Krikun, Shazeer, and Chen]{gshard}
Dmitry Lepikhin, HyoukJoong Lee, Yuanzhong Xu, Dehao Chen, Orhan Firat, Yanping Huang, Maxim Krikun, Noam~M. Shazeer, and Z.~Chen.
\newblock Gshard: Scaling giant models with conditional computation and automatic sharding.
\newblock \emph{ArXiv}, abs/2006.16668, 2020.

\bibitem[Li et~al.(2024)Li, Hammoud, Itani, Khizbullin, and Ghanem]{li2024camel}
Guohao Li, Hasan Hammoud, Hani Itani, Dmitrii Khizbullin, and Bernard Ghanem.
\newblock Camel: Communicative agents for" mind" exploration of large language model society.
\newblock \emph{Advances in Neural Information Processing Systems}, 36, 2024.

\bibitem[Liang et~al.(2022)Liang, Fan, Sarkar, Jiang, Chen, Zou, Cheng, Hao, and Wang]{Liang2022M3ViTMV}
Hanxue Liang, Zhiwen Fan, Rishov Sarkar, Ziyu Jiang, Tianlong Chen, Kai Zou, Yu~Cheng, Cong Hao, and Zhangyang Wang.
\newblock M3vit: Mixture-of-experts vision transformer for efficient multi-task learning with model-accelerator co-design.
\newblock \emph{ArXiv}, abs/2210.14793, 2022.

\bibitem[Liang et~al.(2023)Liang, He, Jiao, Wang, Wang, Wang, Yang, Tu, and Shi]{liang2023encouraging}
Tian Liang, Zhiwei He, Wenxiang Jiao, Xing Wang, Yan Wang, Rui Wang, Yujiu Yang, Zhaopeng Tu, and Shuming Shi.
\newblock Encouraging divergent thinking in large language models through multi-agent debate.
\newblock \emph{arXiv preprint arXiv:2305.19118}, 2023.

\bibitem[Liu et~al.(2023)Liu, Qiao, Neiswanger, Wang, Tan, Tao, Li, Wang, Sun, Pangarkar, Fan, Gu, Miller, Zhuang, He, Li, Koto, Tang, Ranjan, Shen, Ren, Iriondo, Mu, Hu, Schulze, Nakov, Baldwin, and Xing]{liu2023llm360}
Zhengzhong Liu, Aurick Qiao, Willie Neiswanger, Hongyi Wang, Bowen Tan, Tianhua Tao, Junbo Li, Yuqi Wang, Suqi Sun, Omkar Pangarkar, Richard Fan, Yi~Gu, Victor Miller, Yonghao Zhuang, Guowei He, Haonan Li, Fajri Koto, Liping Tang, Nikhil Ranjan, Zhiqiang Shen, Xuguang Ren, Roberto Iriondo, Cun Mu, Zhiting Hu, Mark Schulze, Preslav Nakov, Tim Baldwin, and Eric~P. Xing.
\newblock Llm360: Towards fully transparent open-source llms, 2023.

\bibitem[Matena and Raffel(2022)]{matena2022merging}
Michael Matena and Colin Raffel.
\newblock Merging models with fisher-weighted averaging, 2022.

\bibitem[Muennighoff et~al.(2023)Muennighoff, Liu, Zebaze, Zheng, Hui, Zhuo, Singh, Tang, von Werra, and Longpre]{muennighoff2023octopack}
Niklas Muennighoff, Qian Liu, Armel Zebaze, Qinkai Zheng, Binyuan Hui, Terry~Yue Zhuo, Swayam Singh, Xiangru Tang, Leandro von Werra, and Shayne Longpre.
\newblock Octopack: Instruction tuning code large language models.
\newblock \emph{arXiv preprint arXiv:2308.07124}, 2023.

\bibitem[Mukherjee et~al.(2023)Mukherjee, Mitra, Jawahar, Agarwal, Palangi, and Awadallah]{mukherjee2023orca}
Subhabrata Mukherjee, Arindam Mitra, Ganesh Jawahar, Sahaj Agarwal, Hamid Palangi, and Ahmed Awadallah.
\newblock Orca: Progressive learning from complex explanation traces of gpt-4, 2023.

\bibitem[Nagarajan and Kolter(2021)]{nagarajan2021uniform}
Vaishnavh Nagarajan and J.~Zico Kolter.
\newblock Uniform convergence may be unable to explain generalization in deep learning, 2021.

\bibitem[OpenAI(2023)]{Achiam2023GPT4TR}
OpenAI.
\newblock {GPT-4} technical report.
\newblock volume abs/2303.08774, 2023.

\bibitem[Polo et~al.(2024)Polo, Weber, Choshen, Sun, Xu, and Yurochkin]{polo2024tinybenchmarks}
Felipe~Maia Polo, Lucas Weber, Leshem Choshen, Yuekai Sun, Gongjun Xu, and Mikhail Yurochkin.
\newblock tinybenchmarks: evaluating llms with fewer examples.
\newblock \emph{ArXiv}, abs/2402.14992, 2024.

\bibitem[Rajbhandari et~al.(2022)Rajbhandari, Li, Yao, Zhang, Aminabadi, Awan, Rasley, and He]{rajbhandari2022deepspeedmoe}
Samyam Rajbhandari, Conglong Li, Zhewei Yao, Minjia Zhang, Reza~Yazdani Aminabadi, Ammar~Ahmad Awan, Jeff Rasley, and Yuxiong He.
\newblock Deepspeed-moe: Advancing mixture-of-experts inference and training to power next-generation ai scale, 2022.

\bibitem[Reddi et~al.(2023)Reddi, Miryoosefi, Karp, Krishnan, Kale, Kim, and Kumar]{reddi2023efficient}
Sashank~J Reddi, Sobhan Miryoosefi, Stefani Karp, Shankar Krishnan, Satyen Kale, Seungyeon Kim, and Sanjiv Kumar.
\newblock Efficient training of language models using few-shot learning.
\newblock In \emph{International Conference on Machine Learning}, pages 14553--14568. PMLR, 2023.

\bibitem[Rozi{\`e}re et~al.(2023)Rozi{\`e}re, Gehring, Gloeckle, Sootla, Gat, Tan, Adi, Liu, Remez, Rapin, Kozhevnikov, Evtimov, Bitton, Bhatt, Ferrer, Grattafiori, Xiong, D'efossez, Copet, Azhar, Touvron, Martin, Usunier, Scialom, and Synnaeve]{Rozire2023CodeLO}
Baptiste Rozi{\`e}re, Jonas Gehring, Fabian Gloeckle, Sten Sootla, Itai Gat, Xiaoqing Tan, Yossi Adi, Jingyu Liu, Tal Remez, J{\'e}r{\'e}my Rapin, Artyom Kozhevnikov, I.~Evtimov, Joanna Bitton, Manish~P Bhatt, Cristian~Cant{\'o}n Ferrer, Aaron Grattafiori, Wenhan Xiong, Alexandre D'efossez, Jade Copet, Faisal Azhar, Hugo Touvron, Louis Martin, Nicolas Usunier, Thomas Scialom, and Gabriel Synnaeve.
\newblock Code llama: Open foundation models for code.
\newblock \emph{ArXiv}, abs/2308.12950, 2023.

\bibitem[Rozière et~al.(2024)Rozière, Gehring, Gloeckle, Sootla, Gat, Tan, Adi, Liu, Sauvestre, Remez, Rapin, Kozhevnikov, Evtimov, Bitton, Bhatt, Ferrer, Grattafiori, Xiong, Défossez, Copet, Azhar, Touvron, Martin, Usunier, Scialom, and Synnaeve]{rozière2024code}
Baptiste Rozière, Jonas Gehring, Fabian Gloeckle, Sten Sootla, Itai Gat, Xiaoqing~Ellen Tan, Yossi Adi, Jingyu Liu, Romain Sauvestre, Tal Remez, Jérémy Rapin, Artyom Kozhevnikov, Ivan Evtimov, Joanna Bitton, Manish Bhatt, Cristian~Canton Ferrer, Aaron Grattafiori, Wenhan Xiong, Alexandre Défossez, Jade Copet, Faisal Azhar, Hugo Touvron, Louis Martin, Nicolas Usunier, Thomas Scialom, and Gabriel Synnaeve.
\newblock Code llama: Open foundation models for code, 2024.

\bibitem[Sakaguchi et~al.(2019)Sakaguchi, Bras, Bhagavatula, and Choi]{Sakaguchi2019AnAW}
Keisuke Sakaguchi, Ronan~Le Bras, Chandra Bhagavatula, and Yejin Choi.
\newblock An adversarial winograd schema challenge at scale.
\newblock 2019.

\bibitem[Semnani et~al.(2023)Semnani, Yao, Zhang, and Lam]{semnani-etal-2023-wikichat}
Sina Semnani, Violet Yao, Heidi Zhang, and Monica Lam.
\newblock {W}iki{C}hat: Stopping the hallucination of large language model chatbots by few-shot grounding on {W}ikipedia.
\newblock In Houda Bouamor, Juan Pino, and Kalika Bali, editors, \emph{Findings of the Association for Computational Linguistics: EMNLP 2023}, pages 2387--2413, Singapore, December 2023. Association for Computational Linguistics.
\newblock \doi{10.18653/v1/2023.findings-emnlp.157}.
\newblock URL \url{https://aclanthology.org/2023.findings-emnlp.157}.

\bibitem[Shazeer et~al.(2017)Shazeer, Mirhoseini, Maziarz, Davis, Le, Hinton, and Dean]{shazeer2017outrageously}
Noam Shazeer, Azalia Mirhoseini, Krzysztof Maziarz, Andy Davis, Quoc Le, Geoffrey Hinton, and Jeff Dean.
\newblock Outrageously large neural networks: The sparsely-gated mixture-of-experts layer, 2017.

\bibitem[Shen et~al.(2023)Shen, Zhang, Cao, Tan, Chen, and Gan]{Shen2023ModuleFormerLM}
Yikang Shen, Zheyu Zhang, Tianyou Cao, Shawn Tan, Zhenfang Chen, and Chuang Gan.
\newblock Moduleformer: Learning modular large language models from uncurated data.
\newblock \emph{ArXiv}, abs/2306.04640, 2023.

\bibitem[Shoemake(1985)]{10.1145/325334.325242}
Ken Shoemake.
\newblock Animating rotation with quaternion curves.
\newblock In \emph{Proceedings of the 12th Annual Conference on Computer Graphics and Interactive Techniques}, SIGGRAPH '85, page 245–254, New York, NY, USA, 1985. Association for Computing Machinery.
\newblock ISBN 0897911660.

\bibitem[Sukhbaatar et~al.(2024)Sukhbaatar, Golovneva, Sharma, Xu, Lin, Rozière, Kahn, Li, tau Yih, Weston, and Li]{sukhbaatar2024branchtrainmix}
Sainbayar Sukhbaatar, Olga Golovneva, Vasu Sharma, Hu~Xu, Xi~Victoria Lin, Baptiste Rozière, Jacob Kahn, Daniel Li, Wen tau Yih, Jason Weston, and Xian Li.
\newblock Branch-train-mix: Mixing expert llms into a mixture-of-experts llm, 2024.

\bibitem[Tissera(2023)]{Synthia-7B-v1.2}
Migel Tissera.
\newblock Synthia-70b-v1.2b: Synthetic intelligent agent.
\newblock \url{https://huggingface.co/migtissera/Synthia-13B}, 2023.

\bibitem[Touvron et~al.(2023)Touvron, Martin, Stone, Albert, Almahairi, Babaei, Bashlykov, Batra, Bhargava, Bhosale, Bikel, Blecher, Ferrer, Chen, Cucurull, Esiobu, Fernandes, Fu, Fu, Fuller, Gao, Goswami, Goyal, Hartshorn, Hosseini, Hou, Inan, Kardas, Kerkez, Khabsa, Kloumann, Korenev, Koura, Lachaux, Lavril, Lee, Liskovich, Lu, Mao, Martinet, Mihaylov, Mishra, Molybog, Nie, Poulton, Reizenstein, Rungta, Saladi, Schelten, Silva, Smith, Subramanian, Tan, Tang, Taylor, Williams, Kuan, Xu, Yan, Zarov, Zhang, Fan, Kambadur, Narang, Rodriguez, Stojnic, Edunov, and Scialom]{Touvron2023Llama2O}
Hugo Touvron, Louis Martin, Kevin~R. Stone, Peter Albert, Amjad Almahairi, Yasmine Babaei, Nikolay Bashlykov, Soumya Batra, Prajjwal Bhargava, Shruti Bhosale, Daniel~M. Bikel, Lukas Blecher, Cristian~Cant{\'o}n Ferrer, Moya Chen, Guillem Cucurull, David Esiobu, Jude Fernandes, Jeremy Fu, Wenyin Fu, Brian Fuller, Cynthia Gao, Vedanuj Goswami, Naman Goyal, Anthony~S. Hartshorn, Saghar Hosseini, Rui Hou, Hakan Inan, Marcin Kardas, Viktor Kerkez, Madian Khabsa, Isabel~M. Kloumann, A.~V. Korenev, Punit~Singh Koura, Marie-Anne Lachaux, Thibaut Lavril, Jenya Lee, Diana Liskovich, Yinghai Lu, Yuning Mao, Xavier Martinet, Todor Mihaylov, Pushkar Mishra, Igor Molybog, Yixin Nie, Andrew Poulton, Jeremy Reizenstein, Rashi Rungta, Kalyan Saladi, Alan Schelten, Ruan Silva, Eric~Michael Smith, R.~Subramanian, Xia Tan, Binh Tang, Ross Taylor, Adina Williams, Jian~Xiang Kuan, Puxin Xu, Zhengxu Yan, Iliyan Zarov, Yuchen Zhang, Angela Fan, Melanie Kambadur, Sharan Narang, Aurelien Rodriguez, Robert Stojnic, Sergey Edunov, and
  Thomas Scialom.
\newblock Llama 2: Open foundation and fine-tuned chat models.
\newblock \emph{ArXiv}, abs/2307.09288, 2023.

\bibitem[Verma and Elbayad(2024)]{verma2024merging}
Neha Verma and Maha Elbayad.
\newblock Merging text transformer models from different initializations.
\newblock \emph{arXiv preprint arXiv:2403.00986}, 2024.

\bibitem[Wan et~al.(2024)Wan, Huang, Cai, Quan, Bi, and Shi]{wan2024knowledge}
Fanqi Wan, Xinting Huang, Deng Cai, Xiaojun Quan, Wei Bi, and Shuming Shi.
\newblock Knowledge fusion of large language models.
\newblock In \emph{The Twelfth International Conference on Learning Representations}, 2024.

\bibitem[Wang et~al.(2023{\natexlab{a}})Wang, Polo, Sun, Kundu, Xing, and Yurochkin]{wang2023fusing}
Hongyi Wang, Felipe~Maia Polo, Yuekai Sun, Souvik Kundu, Eric Xing, and Mikhail Yurochkin.
\newblock Fusing models with complementary expertise, 2023{\natexlab{a}}.

\bibitem[Wang et~al.(2023{\natexlab{b}})Wang, Panda, Hennigen, Greengard, Karlinsky, Feris, Cox, Wang, and Kim]{wang2023learning}
Peihao Wang, Rameswar Panda, Lucas~Torroba Hennigen, Philip Greengard, Leonid Karlinsky, Rogerio Feris, David~Daniel Cox, Zhangyang Wang, and Yoon Kim.
\newblock Learning to grow pretrained models for efficient transformer training.
\newblock \emph{arXiv preprint arXiv:2303.00980}, 2023{\natexlab{b}}.

\bibitem[Wang et~al.(2023{\natexlab{c}})Wang, Panda, and Wang]{wang2023data}
Peihao Wang, Rameswar Panda, and Zhangyang Wang.
\newblock Data efficient neural scaling law via model reusing.
\newblock In \emph{International Conference on Machine Learning}, pages 36193--36204. PMLR, 2023{\natexlab{c}}.

\bibitem[Wei et~al.(2023)Wei, Wang, Liu, Ding, and Zhang]{magicoder}
Yuxiang Wei, Zhe Wang, Jiawei Liu, Yifeng Ding, and Lingming Zhang.
\newblock Magicoder: Source code is all you need, 2023.

\bibitem[Wortsman et~al.(2022)Wortsman, Ilharco, Gadre, Roelofs, Gontijo-Lopes, Morcos, Namkoong, Farhadi, Carmon, Kornblith, and Schmidt]{wortsman2022model}
Mitchell Wortsman, Gabriel Ilharco, Samir~Yitzhak Gadre, Rebecca Roelofs, Raphael Gontijo-Lopes, Ari~S. Morcos, Hongseok Namkoong, Ali Farhadi, Yair Carmon, Simon Kornblith, and Ludwig Schmidt.
\newblock Model soups: averaging weights of multiple fine-tuned models improves accuracy without increasing inference time, 2022.

\bibitem[Wu et~al.(2024)Wu, Gan, Ge, Lu, Wang, Feng, Luo, and Shan]{wu2024llama}
Chengyue Wu, Yukang Gan, Yixiao Ge, Zeyu Lu, Jiahao Wang, Ye~Feng, Ping Luo, and Ying Shan.
\newblock Llama pro: Progressive llama with block expansion.
\newblock \emph{arXiv preprint arXiv:2401.02415}, 2024.

\bibitem[Wu et~al.(2023)Wu, Bansal, Zhang, Wu, Zhang, Zhu, Li, Jiang, Zhang, and Wang]{wu2023autogen}
Qingyun Wu, Gagan Bansal, Jieyu Zhang, Yiran Wu, Shaokun Zhang, Erkang Zhu, Beibin Li, Li~Jiang, Xiaoyun Zhang, and Chi Wang.
\newblock Autogen: Enabling next-gen llm applications via multi-agent conversation framework.
\newblock \emph{arXiv preprint arXiv:2308.08155}, 2023.

\bibitem[Xu et~al.(2024)Xu, Yuan, Wang, Wang, Song, and Song]{xu2024training}
Zhengqi Xu, Ke~Yuan, Huiqiong Wang, Yong Wang, Mingli Song, and Jie Song.
\newblock Training-free pretrained model merging.
\newblock \emph{arXiv preprint arXiv:2403.01753}, 2024.

\bibitem[Yadav et~al.(2023)Yadav, Tam, Choshen, Raffel, and Bansal]{yadav2023tiesmerging}
Prateek Yadav, Derek Tam, Leshem Choshen, Colin Raffel, and Mohit Bansal.
\newblock Ties-merging: Resolving interference when merging models, 2023.

\bibitem[Yu et~al.(2024)Yu, Yu, Yu, Huang, and Li]{yu2024language}
Le~Yu, Bowen Yu, Haiyang Yu, Fei Huang, and Yongbin Li.
\newblock Language models are super mario: Absorbing abilities from homologous models as a free lunch, 2024.

\bibitem[Yu et~al.(2023)Yu, Jiang, Shi, Yu, Liu, Zhang, Kwok, Li, Weller, and Liu]{yu2023metamath}
Longhui Yu, Weisen Jiang, Han Shi, Jincheng Yu, Zhengying Liu, Yu~Zhang, James~T Kwok, Zhenguo Li, Adrian Weller, and Weiyang Liu.
\newblock Metamath: Bootstrap your own mathematical questions for large language models.
\newblock \emph{arXiv preprint arXiv:2309.12284}, 2023.

\bibitem[Zheng et~al.(2023)Zheng, Chiang, Sheng, Zhuang, Wu, Zhuang, Lin, Li, Li, Xing, Zhang, Gonzalez, and Stoica]{zheng2023judging}
Lianmin Zheng, Wei-Lin Chiang, Ying Sheng, Siyuan Zhuang, Zhanghao Wu, Yonghao Zhuang, Zi~Lin, Zhuohan Li, Dacheng Li, Eric~P. Xing, Hao Zhang, Joseph~E. Gonzalez, and Ion Stoica.
\newblock Judging llm-as-a-judge with mt-bench and chatbot arena, 2023.

\bibitem[Zheng et~al.(2024)Zheng, Chiang, Sheng, Zhuang, Wu, Zhuang, Lin, Li, Li, Xing, et~al.]{zheng2024judging}
Lianmin Zheng, Wei-Lin Chiang, Ying Sheng, Siyuan Zhuang, Zhanghao Wu, Yonghao Zhuang, Zi~Lin, Zhuohan Li, Dacheng Li, Eric Xing, et~al.
\newblock Judging llm-as-a-judge with mt-bench and chatbot arena.
\newblock \emph{Advances in Neural Information Processing Systems}, 36, 2024.

\bibitem[Zoph et~al.(2022)Zoph, Bello, Kumar, Du, Huang, Dean, Shazeer, and Fedus]{Zoph2022DesigningES}
Barret Zoph, Irwan Bello, Sameer Kumar, Nan Du, Yanping Huang, Jeff Dean, Noam~M. Shazeer, and William Fedus.
\newblock Designing effective sparse expert models.
\newblock \emph{2022 IEEE International Parallel and Distributed Processing Symposium Workshops (IPDPSW)}, pages 1044--1044, 2022.

\end{thebibliography}

\newpage
\section*{Checklist}

\vspace{-2mm}
\begin{enumerate}

\item For all authors...
\begin{enumerate}
\vspace{-2mm}
  \item Do the main claims made in the abstract and introduction accurately reflect the paper's contributions and scope?
    \answerYes{Our abstract and introduction in Section~\ref{sec:introduction} accurately reflect the paper's contributions and scope?.}
    \vspace{-2mm}
  \item Did you describe the limitations of your work?
    \answerYes{In Section~\ref{sec6} and~\ref{conclusion} we discuss two other LLM scaling methods for future work and present some preliminary results. In Appendix~\ref{sec:appendix} we discuss its limitations.}
    \vspace{-2mm}
  \item Did you discuss any potential negative societal impacts of your work?
    \answerNA{Our work focuses on foundational research and is not directly related to societal impacts.}
    \vspace{-2mm}
  \item Have you read the ethics review guidelines and ensured that your paper conforms to them?
    \answerYes{We have read and follow the ethics review guidelines.}
\end{enumerate}
\vspace{-2mm}
\item If you are including theoretical results...
\vspace{-2mm}
\begin{enumerate}
  \item Did you state the full set of assumptions of all theoretical results?
    \answerNA{Our work does not include theoretical results.}
    \vspace{-2mm}
  \item Did you include complete proofs of all theoretical results?
    \answerNA{Our work does not include theoretical results.}
\end{enumerate}
\vspace{-2mm}
\item If you ran experiments (e.g. for benchmarks)...
\vspace{-2mm}
\begin{enumerate}
  \item Did you include the code, data, and instructions needed to reproduce the main experimental results (either in the supplemental material or as a URL)?
    \answerYes{We have provided our code, data (the way to get them), and instructions needed for reproducing in a GitHub repository.}
    \vspace{-2mm}
  \item Did you specify all the training details (e.g., data splits, hyperparameters, how they were chosen)?
    \answerYes{We have provided all the training details in our Appendix.}
    \vspace{-2mm}
	\item Did you report error bars (e.g., with respect to the random seed after running experiments multiple times)?
    \answerNA{Our main results do not contain randomness, and running multiple times with different random seeds leads to the same results.}
    \vspace{-2mm}
	\item Did you include the total amount of compute and the type of resources used (e.g., type of GPUs, internal cluster, or cloud provider)?
    \answerYes{We have provided the computing resources we used in our Appendix.}
\end{enumerate}
\vspace{-2mm}
\item If you are using existing assets (e.g., code, data, models) or curating/releasing new assets...
\begin{enumerate}
\vspace{-2mm}
  \item If your work uses existing assets, did you cite the creators?
    \answerYes{We have cited them.}
  \item Did you mention the license of the assets?
    \answerYes{We have mentioned the license of all the datasets we used in Appendix~\ref{sec:appendix}.}
  \vspace{-2mm}
  \item Did you include any new assets either in the supplemental material or as a URL?
    \answerNA{We do not release any new assets.}
 \vspace{-2mm}
  \item Did you discuss whether and how consent was obtained from people whose data you're using/curating?
    \answerNA{We only use public available data in this work.}
 \vspace{-2mm}
  \item Did you discuss whether the data you are using/curating contains personally identifiable information or offensive content?
    \answerNA{The data we are using/curating does not contain any of these.}
\end{enumerate}
\vspace{-2mm}
\item If you used crowdsourcing or conducted research with human subjects...
\vspace{-2mm}
\begin{enumerate}
  \item Did you include the full text of instructions given to participants and screenshots, if applicable?
    \answerNA{We did not use any crowdsourcing or conduct research with human subjects.}
    \vspace{-2mm}
  \item Did you describe any potential participant risks, with links to Institutional Review Board (IRB) approvals, if applicable?
    \answerNA{We did not use any crowdsourcing or conduct research with human subjects.}
    \vspace{-2mm}
  \item Did you include the estimated hourly wage paid to participants and the total amount spent on participant compensation?
    \answerNA{We did not use any crowdsourcing or conduct research with human subjects.}
    \vspace{-2mm}
\end{enumerate}

\end{enumerate}

\newpage
\appendix

\section*{Appendix}\label{sec:appendix}

\section{Implementation Details}

\subsection{Detailed Algorithms of Heuristic Strategy of Model Merging}\label{merge_algo}

\paragraph{Heuristic (Average).} We present the implementation details in Algorithm \ref{alg:heuristicaverage}. The algorithm takes a mergable model family as input and generate a merged model as output. For each candidate model in input model family, we compute the accuracy of the temporary merged model, generated by the union of this candidate model and the previously selected model, on the proxy dataset, and the candidate that brings no harm to the accuracy will be selected for the final merged model. Each weight of the merged model is generated by averaging the corresponding weights of all the selected models.

\begin{algorithm}
\caption{Heuristic (Average)}
\label{alg:heuristicaverage}
\renewcommand{\algorithmicrequire}{\textbf{Input:}}
\renewcommand{\algorithmicensure}{\textbf{Output:}}
\begin{algorithmic}[1]
\Require A mergable family $\{w_{1}, ..., w_{n}\}$ (sorted in decreasing order of Acc($w_{i}$)).
\Ensure \textit{merged\_model}
\State \textit{models\_to\_merge} $\gets \{w_{1}\}$
\State \textit{merged\_model} $\gets w_{1}$
\For{$i = 2$ to $n$}
    \If{ProxyAcc(AvgMerge(\textit{models\_to\_merge}  $\cup \{w_{i}\})) \geq$ ProxyAcc(\textit{merged\_model}))}
        \State \textit{models\_to\_merge} $\gets$ \textit{models\_to\_merge} $\cup \{w_{i}\}$
        \State \textit{merged\_model} = AvgMerge(\textit{models\_to\_merge})

    \EndIf
\EndFor
\State \Return \textit{merged\_model}
\end{algorithmic}
\end{algorithm}

\paragraph{Heuristic (Coefficient).} We present the implementation details in Algorithm \ref{alg:heuristiccoefficient}. Heuristic (Coefficient) builds upon Heuristic (Average) by combining the previously merged model with a new candidate using different coefficients in each round. To reduce the search space, we set the range of coefficient as {0.1, 0.2...0.9}. 

\begin{algorithm}
\caption{Heuristic (Coefficient)}
\label{alg:heuristiccoefficient}
\renewcommand{\algorithmicrequire}{\textbf{Input:}}
\renewcommand{\algorithmicensure}{\textbf{Output:}}
\begin{algorithmic}[1]
\Require A mergable family $\{w_{1}, ..., w_{n}\}$  (sorted in decreasing order of Acc($w_{i}$)), a list of coefficients $\{0.1, 0.2 ..., 0.9\}$ to be searched when merging.
\Ensure merged\_model
\State \textit{coefficients} $\gets \{0.1, 0.2 ..., 0.9\}$
\State \textit{merged\_model} $\gets w_{1}$

\For{$i = 2$ to $n$}
    \State \textit{best\_acc}, \textit{best\_c} $\gets$ \text{ProxyAcc}(\textit{merged\_model}), $1.0$
    \For{$c$ in \textit{coefficients}}
        \If{ProxyAcc(Merge($c$, \textit{merged\_model}, $w_{i}$)) $\geq$ \textit{best\_acc}}
            \State \textit{best\_acc}, \textit{best\_c} $\gets$ ProxyAcc(Merge($c$, \textit{merged\_model}, $w_{i}$)) , $c$
        \EndIf
    \EndFor
    \State \textit{merged\_model} $\gets$ Merge(\textit{best\_c}, \textit{merged\_model}, $w_{i}$)
\EndFor
\State \Return \textit{merged\_model}
\end{algorithmic}
\end{algorithm}

\paragraph{Heuristic (Similarity).} We present the implementation details in Algorithm \ref{alg:heuristicsimilarity}. We use the average similarity of all weights as the criterion for selecting models in each round. This algorithm selects the candidate model with the highest or lowest similarity and conducts a conefficient search to combine it with the previously merged model.

\newpage
\begin{algorithm}
\caption{Heuristic (Similarity)}
\label{alg:heuristicsimilarity}
\renewcommand{\algorithmicrequire}{\textbf{Input:}}
\renewcommand{\algorithmicensure}{\textbf{Output:}}
\begin{algorithmic}[1]
\Require A mergable family $\{w_{1}, ..., w_{n}\}$ (sorted in decreasing order of Acc($w_{i}$)), a list of coefficients $\{0.1, 0.2 ..., 0.9\}$ to be searched when merging.
\Ensure merged\_model
\State \textit{merged\_model} $\gets w_{1}$
\State \textit{remaining\_models} $\gets \{w_{2}, ..., w_{n}\}$
\For{$i = 2$ to $n$}
    \State \textit{best\_acc}, \textit{best\_c} $\gets$ \text{ProxyAcc}(\textit{merged\_model}), $1.0$
    \State \textit{candidate\_model} $\gets$ GetModelBySimilarity(\textit{merged\_model}, \textit{remaining\_models})
    \For{$c$ in \textit{coefficients}}
        \If{ProxyAcc(Merge($c$, \textit{merged\_model}, \textit{candidate\_model})) $\geq$ \textit{best\_acc}}
            \State \textit{best\_acc}, \textit{best\_c} $\gets$ ProxyAcc(Merge($c$, \textit{merged\_model}, \textit{candidate\_model})) , $c$
        \EndIf
    \EndFor
    \State \textit{merged\_model} $\gets$ Merge(\textit{best\_c}, \textit{merged\_model}, \textit{candidate\_model})
    \State \textit{remaining\_models} $\gets $\textit{remaining\_models} $ \setminus \{\text{\textit{candidate\_model}}\}$
\EndFor
\State \Return \textit{merged\_model}
\end{algorithmic}
\end{algorithm}

\subsection{Detailed about Evolutionary Strategy of Model Merging}\label{apdx:details_of_evo}
For the experiments of \textbf{Q2} - (\textit{i}) in Section~\ref{Merging_Bench}, we constrain all parameter values to be within the range of $[0, 1]$. TIES and DARE require to optimize $2*k$ parameters, while other methods require to optimize $k$ parameters, where $k$ represents the number of models included in the model zoo.

For the experiments of \textbf{Q2} - (\textit{ii}) in Section~\ref{Merging_Bench}, we choose the Linear method for experimentation, and we constrain all parameter values to be within the range of $[0, 1]$. 
For finer-grained merging, we group adjacent $n$ decoder layers together, where they share the same coefficient. For the remaining parameters, we make them share the same coefficient. Hence, the number of parameters that need to be fine-tuned is given by:
$k*(\frac{\text{\textit{num\_hidden\_layers}}}{n}+1)$, where $k$ represents the number of models and $n$ represents the size of groups. For the case of $n=32$, we utilized the previous results, thus the number of parameters to be optimized is $k$.

For the experiments of \textbf{Q2} - (\textit{iii}) in Section~\ref{Merging_Bench}, we control the variation of coefficients obtained through heuristic strategy to not exceed $0.1$, and when it is negative, we set it to $0$. We also only evaluate the Linear method.

\subsection{Detailed Algorithms of Model Mixture}\label{apdx:algo_mixture}

\paragraph{Model Level Mixture.} We present the implementation details in Algorithm \ref{alg:model_level_mixture}. The mixed model consists of a router, which determines the expert to execute inference, and all the input models as experts. All the weights of input model's components, including embedding layers (embd\_layer), decoder layers (layers) and language model head (lm\_head), will be integrated into the mixed model.

\begin{algorithm}
\caption{Model Level Mixture}
\label{alg:model_level_mixture}
\renewcommand{\algorithmicrequire}{\textbf{Input:}}
\renewcommand{\algorithmicensure}{\textbf{Output:}}
\begin{algorithmic}[1]
\Require A model family $\{w_{1}, ..., w_{n}\}$
\Ensure \textit{mixed\_model}
\State \textit{mixed\_model.router} $\gets$ GenerateRouter($\{w_{1}, ..., w_{n}\}$)
\For{$i = 1$ to $n$}
    \State \textit{mixed\_model.$expert_{i}$} $\gets$ $w_{i}$
\EndFor
\State \Return \textit{mixed\_model}
\end{algorithmic}
\end{algorithm}

\paragraph{Block Level Mixture.} We present the implementation details in Algorithm \ref{alg:block_level_mixture}. Different from model-level mixture, block-level mixture utilizes the embd\_layer and lm\_head of an additional model within a model family to handle input and output. Meanwhile, the transformer blocks of other models within the model family act as experts, connected by a router.

\begin{algorithm}
\caption{Block Level Mixture}
\label{alg:block_level_mixture}
\renewcommand{\algorithmicrequire}{\textbf{Input:}}
\renewcommand{\algorithmicensure}{\textbf{Output:}}
\begin{algorithmic}[1]
\Require A model family $\{w_{1}, ..., w_{n}\}$ with identical layer amount, one of the family as $base\_model$
\Ensure \textit{mixed\_model}
\State \textit{mixed\_model.embd\_layer} $\gets$ \textit{base\_model.embd\_layer}
\State \textit{mixed\_model.lm\_head} $\gets$ \textit{base\_model.lm \_head}
\For{$i = 0$ to Len(\textit{base\_model.layers})}
    \State \textit{mixed\_model.$layer_{i}$.router} $\gets$ GenerateRouter($\{w_{1}, ..., w_{n}\}$)
    \For{$j=1$ to $n$}
        \State \textit{mixed\_model.$layer_{i}.expert_{j}$} $\gets$ $w_{j}.layer_{i}$
    \EndFor
\EndFor
\State \Return \textit{mixed\_model}
\end{algorithmic}
\end{algorithm}

\paragraph{FFN Level Mixture.} We present the implementation details in Algorithm \ref{alg:ffn_level_mixture}. FFN level mixture is similar to block level with only difference on inner-block component sharing. Each layer of the mixed model will take the attention weights of the base model and build an MoE structure based on the FFNs in corresponding layers of all the input models.

\begin{algorithm}
\caption{FFN Level Mixture}
\label{alg:ffn_level_mixture}
\renewcommand{\algorithmicrequire}{\textbf{Input:}}
\renewcommand{\algorithmicensure}{\textbf{Output:}}
\begin{algorithmic}[1]
\Require A model family $\{w_{1}, ..., w_{n}\}$ with identical layer amount, one of the family as $base\_model$.
\Ensure \textit{mixed\_model}, 
\State \textit{mixed\_model.embd\_layer} $\gets$ \textit{base\_model.embd\_layer}
\State \textit{mixed\_model.lm\_head} $\gets$ \textit{base\_model.lm\_head}
\For{$i = 0$ to Len(\textit{base\_model.layers})}
    \State \textit{mixed\_model.$layer_{i}$.router} $\gets$ GenerateRouter($\{w_{1}, ..., w_{n}\}$)
    \State \textit{mixed\_model.$layer_{i}$.attention} $\gets$ \textit{base\_model.$layer_{i}$.attention}
    \State \textit{mixed\_model.$layer_{i}$.norm} $\gets$ \textit{base\_model.$layer_{i}$.norm}
    \For{$j=1$ to $n$}
        \State \textit{mixed\_model.$layer_{i}.expert_{j}$} $\gets$ \textit{$w_{j}.layer_{i}$.FFN}
    \EndFor
\EndFor
\State \Return \textit{mixed\_model}
\end{algorithmic}
\end{algorithm}

\paragraph{Hybrid Mixture} We present the implementation details in Algorithm \ref{alg:hybrid_mixture}. The hybrid mixture combines both merging and mixture methods. Specifically, the first $k$ layers of the mixed model are obtained by merging multiple models, while the rest of the layers use an FFN-level mixture architecture.

\begin{algorithm}
\caption{Hybrid Mixture}
\label{alg:hybrid_mixture}
\renewcommand{\algorithmicrequire}{\textbf{Input:}}
\renewcommand{\algorithmicensure}{\textbf{Output:}}
\begin{algorithmic}[1]
\Require A model family $\{w_{1}, ..., w_{n}\}$ with identical layer amount, one of the family as $base\_model$, $k$ layers for merging and the rest layers for mixture.
\Ensure \textit{mixed\_model}
\State \textit{mixed\_model.embd\_layer} $\gets$ \textit{base\_model.embd\_layer}
\State \textit{mixed\_model.lm\_head} $\gets$ \textit{base\_model.lm\_head}
\For{$i=0$ to $k$}
    \State \textit{mixed\_model.$layer_{i}$} $\gets$ Merge($\{w_{1}, ..., w_{n}\}$, i)
\EndFor
\For{$i = k + 1$ to Len(\textit{base\_model.layers})}
    \State \textit{mixed\_model.$layer_{i}$.router} $\gets$ GenerateRouter($\{w_{1}, ..., w_{n}\}$)
    \State \textit{mixed\_model.$layer_{i}$.attention} $\gets$ \textit{base\_model.$layer_{i}$.attention}
    \State \textit{mixed\_model.$layer_{i}$.norm} $\gets$ \textit{base\_model.$layer_{i}$.norm}
    \For{$j=1$ to $n$}
        \State \textit{mixed\_model.$layer_{i}.expert_{j}$.FFN} $\gets$ \textit{$w_{j}.layer_{i}$.FFN}
    \EndFor
\EndFor
\State \Return \textit{mixed\_model}
\end{algorithmic}
\end{algorithm}

\subsection{Details of Model-Glue}\label{apdx:modelglue}
The models selected by the heuristic strategy include: migtissera/Synthia-7B-v1.2, neuralmagic/Llama-2-7b-evolcodealpaca, teknium/OpenHermes-7B, meta-llama/Llama-2-7b-chat-hf, meta-math/MetaMath-7B-V1.0, lmsys/vicuna-7b-v1.5.
Since merging ise-uiuc/Magicoder-S-CL-7B and codellama/CodeLlama-7b-Instruct-hf does not lead to improvement in the Codellama's mergeable family, we select ise-uiuc/Magicoder-S-CL-7B as the representative model.

The final models used for Model-level Mixture are: LLM360/CrystalChat, ise-uiuc/Magicoder-S-CL-7B, meta-math/MetaMath-Llemma-7B and the representative model of the Llama-2 family obtained through the Heuristic (Coefficient).
Please refer to our repository for specific configurations.

\subsection{Details of clustering in selective merging pipeline}\label{apdx:clustering}
\paragraph{Motivation for using cosine similarity as a model selection criterion} Previous merging study~\cite{yu2024language} finds that merging performance is consistent with parameter similarity. We inherit it by using cosine similarity as a representative method to measure whether a model can be merged.
From our preliminary result, cosine similarity works effectively. Empirically, when the cosine similarity between models exceeds $0.95$, merging them can yield positive benefits. In Table~\ref{tab:mergability}, we present examples of successful and unsuccessful merging. For example, the cosine similarity between the weights of Llama-2-chat and Vicuna is $0.9982$, resulting in the merged model significantly outperforming its parent models. On the other hand, the cosine similarity between the weights of Llama-2-chat and CodeLlama is $0.5351$, indicating that the merged model is inferior to CodeLlama.
Moreover, using cosine similarity to measure the merging benefit is simple and efficient. For these reasons, we stick with cosine similarity for selective merging pipelines.
\vspace{-2mm}
\paragraph{Criteria for Determining the Number of Clusters.}
We cluster models with cosine similarity greater than $0.95$ into a mergeable family, ensuring that within this mergeable family, the pairwise similarities between models are greater than $0.95$.
The number of clusters is automatically determined during the process, after which we execute our merge strategy within each cluster. For \texttt{Which16} model zoo in our paper, we clustered $16$ models and finally obtained five mergeable families:
\ding{182} 12 models fine-tuned based on llama-2, \ding{183}  ise-uiuc/Magicoder-S-CL-7B, \ding{184} codellama/CodeLlama-7b-Instruct-hf, \ding{185} meta-math/MetaMath-Llemma-7B, \ding{186} LLM360/CrystalChat.
Since the remaining clusters contain only one model each, we only report the results of different merging strategies performed within Family \ding{182}.
\vspace{-2mm}
\paragraph{Impact of clustering threshold}
We computed the cosine similarity between 12 LLMs all fine-tuned from Llama-2. These models are considered to be well mergeable, having the same architecture and initialization. Since their similarities range from $0.9680$ to $0.9999$, $0.95$ could be a lower bound for model clustering.
To show the impact of different clustering thresholds, we have examined the performance of merged models with drastically different similarity: Llama-2,deepseek-coder, CodeLlama, and MetaMath-Llema. We use linear interpolation to merge two models and present the benchmarking results in Table~\ref{tab:clustering_1}. The performance of the individual models is shown in Table~\ref{tab:clustering_2}. If the merged model outperforms its parent models on average accuracy, we consider it a successful merge.
From Table~\ref{tab:clustering_1}, we see that successful merging only occurs between Codellama and Codellama-instruct whose weights reach $0.99$ similarity and have the same initialization. To include more mergeable models, we finally choose $0.95$ as the threshold for clustering.

\begin{table}
\centering
\caption{\small Performance of merged models with different similarity. Sim. stands for cosine similarity.} \label{tab:clustering_1}
\renewcommand\arraystretch{1}
\tabcolsep=0.1cm
\resizebox{\linewidth}{!}{
\begin{tabular}{cc|ccccccc|c}
\toprule
\midrule
Parent Model 1           & Parent Model 2           & ARC     & MMLU    & WinoGrande & GSM8K   & HumanEval & MBPP    & Avg.    & Sim. \\ \midrule
Llama-2-7b-hf            & deepseek-coder-6.7b-base & 27.73\% & 24.38\% & 49.64\%    & 0.00\%  & 0.00\%    & 0.00\%  & 16.96\% & 0\%      \\
Llama-2-7b-hf            & CodeLlama-7b-hf          & 41.04\% & 31.68\% & 66.85\%    & 5.76\%  & 10.98\%   & 21.40\% & 29.62\% & 52.55\%      \\
CodeLlama-7b-Python-hf   & CodeLlama-7b-hf          & 40.61\% & 37.17\% & 65.35\%    & 6.67\%  & 21.95\%   & 25.60\% & 32.89\% & 60.34\%      \\
MetaMath-Llemma-7B       & CodeLlama-7b-hf          & 46.16\% & 42.86\% & 64.64\%    & 27.07\% & 34.76\%   & 37.40\% & 42.15\% & 88.70\%      \\
CodeLlama-7b-Instruct-hf & CodeLlama-7b-hf          & 43.86\% & 41.39\% & 68.59\%    & 16.07\% & 33.54\%   & 40.80\% & 40.71\% & 99.94\%  \\
\midrule
\bottomrule
\end{tabular}}
\end{table}

\begin{table}
\centering
\caption{\small Performance of parent models.} \label{tab:clustering_2}
\renewcommand\arraystretch{1}
\tabcolsep=0.1cm
\resizebox{\linewidth}{!}{
\begin{tabular}{c|cccccccc}
\toprule
\midrule
Model          & ARC & MMLU & WinoGrande & GSM8K & HumanEval & MBPP & Avg. \\ \midrule
Llama-2-7b-hf            & 53.92\%      & 45.83\%       & 74.11\%             & 13.72\%        & 10.98\%            & 18.00\%       & 36.09\%       \\
deepseek-coder-6.7b-base & 36.86\%      & 36.36\%       & 57.30\%             & 19.03\%        & 45.12\%            & 54.80\%       & 41.58\%       \\
CodeLlama-7b-hf          & 41.89\%      & 39.05\%       & 65.98\%             & 11.83\%        & 32.32\%            & 37.20\%       & 38.05\%       \\
CodeLlama-7b-Python-hf   & 40.70\%      & 35.62\%       & 64.56\%             & 13.12\%        & 38.41\%            & 41.20\%       & 38.94\%       \\
MetaMath-Llemma-7B       & 46.67\%      & 46.29\%       & 64.33\%             & 62.24\%        & 32.32\%            & 42.00\%       & 48.97\%       \\
CodeLlama-7b-Instruct-hf & 43.00\%      & 41.69\%       & 65.90\%             & 18.12\%        & 33.70\%            & 40.00\%       & 40.40\%      \\
\midrule
\bottomrule
\end{tabular}}
\end{table}  

\subsection{Energy Consumption}
Existing literature is mainly concerned with carbon emissions during LLM pre-training~\cite{Dodge2022MeasuringTC,Touvron2023Llama2O}. However, the training costs associated with the approaches evaluated in our benchmark are minimal. Specifically, the only training expenditure in our study pertains to the B-M-S router training, as described in Section~\ref{impl_mixture}. This process requires about $80$ GPU hours, resulting in $13.55$kg CO2 emissions based on a $400$W power consumption. In contrast, LLaMA-2-7B pre-training results in $31.22$t CO2, which is over $2000$ times more than ours.

\section{Additional Results}
\subsection{Experiment on Mistral model family}
\label{sec:mistral}
We choose the Llama2-based model family for the main experiments because there are more diverse variances built on different datasets and training recipes. There are many domain-specific models based on Llama-2, such as those for code, mathematics, healthcare, finance, law, and mental health. Importantly, a series of models have undergone continuous pre-training based on Llama-2, and a considerable portion of models trained from scratch have drawn inspiration from the architecture of Llama-2. While these models share the same architecture as Llama-2, their weights exhibit significant differences. Thus, we can thoroughly examine the effect of merging, mixture and Model-GLUE on different settings.
To further evaluate our proposal on the Mistral model family, we have established a Mistral-based \texttt{Which8} model zoo and replicated the experiments outlined in Section~\ref{sec:model_glue}. From the result in Table~\ref{tab:mistral} it can be seen that \texttt{Mode-GLUE} consistently outperform.

\begin{table}
\centering
\caption{\small Comparison between the best single model, Merging, Full Mixture and our Model-GLUE with Mistral model zoo. We highlight the better performance in \textbf{bold}.} \label{tab:mistral}
\renewcommand\arraystretch{1}
\tabcolsep=0.1cm
\resizebox{0.8\linewidth}{!}{
\begin{tabular}{c|ccccccc}
\toprule
\midrule
Model & ARC  & WinoGrande  & MMLU & GSM8K & MBPP & HumanEval & Average \\ 
\midrule
Best Single Model & $\mathbf{67.24}\%$ & $79.01\%$ & $61.77\%$ & $63.15\%$ & $35.98\%$ & $39.00\%$ & $57.69\%$ \\ 
DARE  &  $64.33\%$  &  $78.37\%$  &  $63.27\%$       & $63.31\%$  &  $39.02\%$ & $\mathbf{44.60}\%$  & $58.82\%$   \\ 
TIES & $63.74\%$ & $77.90\%$ & $60.90\%$ & $49.13\%$ & $34.76\%$ & $39.40\%$ & $54.30\%$ \\ 
F-L-S  &  $64.85\%$  &  $\mathbf{79.72\%}$  &  $63.42\%$       & $64.82\%$  &  $42.00\%$ & $42.07\%$  & $59.48\%$   \\
\texttt{Model-GLUE}  &  $65.02\%$  &  $78.85\%$  &  $\mathbf{64.39}\%$       & $65.50\%$  &  $\mathbf{44.60}\%$ & $42.68\%$  & $\mathbf{60.18}\%$   \\
\midrule
\bottomrule
\end{tabular}}
\vspace{-10pt}
\end{table}

\subsection{Model Merging}
\label{sec:more_merging}
We present the specific results of Figure~\ref{fig:merge_radar} in Table~\ref{tab:heuristic} and Table~\ref{tab:evo-methods} and other results of Section~\ref{Merging_Bench} in Table~\ref{tab:mergability}, Table~\ref{tab:evo-group-size} and Table~\ref{tab:efficient-merge}.

\begin{table}[h]
\centering
\caption{\small Failure case of existing merging approaches when expanding the model zoo.\label{tab:mergability}}
\renewcommand\arraystretch{1}
\tabcolsep=0.1cm
\resizebox{0.8\linewidth}{!}{
\begin{tabular}{c|cccccc|c}
\toprule
\midrule
Merging Method & ARC  & WinoGrande  & MMLU & GSM8K & MBPP & HumanEval & Average \\ 
\midrule
\multicolumn{8}{c}{\texttt{Single Model}} \\
\midrule
Llama-2-chat             & 54.10\%          & 71.27\%          & 47.28\%          & 23.05\%          & 17.00\%          & 13.41\%          & 37.68\%          \\
Vicuna                   & 53.75\%          & 70.56\%          & 49.78\%          & 19.11\%          & 6.00\%           & 19.51\%          & 36.45\%          \\
CodeLlama                & 43.52\%          & 65.11\%          & 41.83\%          & 17.06\%          & 40.00\% & 33.70\% & 40.20\%          \\
\midrule
\multicolumn{8}{c}{\texttt{Merge Llama-2-chat and Vicuna}} \\
\midrule
Linear      & 54.27\%          & 72.30\%          & \textbf{50.72\%} & 24.49\% & 20.80\%          & \textbf{20.12\%} & 40.45\%          \\
Model Stock & 54.61\%          & \textbf{74.43\%} & 47.44\%          & 16.07\% & \textbf{22.40\%} & 14.02\%          & 38.16\%          \\
SLERP       & \textbf{55.29\%} & 72.45\%          & 50.51\%          & 24.87\% & 21.80\%          & \textbf{20.12\%} & \textbf{40.84\%} \\
Task Arithmetic & 54.27\% & 71.67\% & 49.95\% & 26.31\%          & 21.40\% & 17.07\% & 40.11\% \\
DARE            & 54.35\% & 72.14\% & 50.38\% & \textbf{26.61\%} & 21.00\% & 17.68\% & 40.36\% \\
TIES            & 52.65\% & 69.93\% & 49.84\% & 24.34\%          & 17.60\% & 19.51\% & 38.98\%\\
\midrule
\multicolumn{8}{c}{\texttt{Merge Llama-2-chat and CodeLlama}} \\
\midrule
Linear          & 45.05\% & 67.09\% & 39.03\% & 16.76\% & \textbf{36.60\%} & \textbf{23.17\%} & 37.95\% \\
Model Stock     & 50.34\% & 71.27\% & 41.06\% & 10.01\% & 15.40\%          & 7.93\%  & 32.67\% \\
SLERP & \textbf{52.05\%} & \textbf{71.43\%} & \textbf{46.41\%} & \textbf{18.95\%} & 20.80\% & 18.90\% & \textbf{38.09\%} \\
Task Arithmetic & 44.97\% & 68.03\% & 38.83\% & 7.05\%  & 10.60\%          & 12.20\% & 30.28\% \\
DARE            & 38.91\% & 65.98\% & 31.90\% & 3.34\%  & 15.00\%          & 9.76\%  & 27.48\% \\
TIES            & 21.67\% & 49.88\% & 25.25\% & 0.00\%  & 0.00\%           & 0.00\%  & 16.13\% \\
\bottomrule
\end{tabular}}
\end{table}

\begin{table}[htb]
\centering
\caption{\small Comparison between different Heuristic Strategies. } \label{tab:heuristic}
\renewcommand\arraystretch{1}
\tabcolsep=0.1cm
\resizebox{0.8\linewidth}{!}{
\begin{tabular}{c|cccccc|c}
\toprule
\midrule
Heuristic Strategy & ARC  & WinoGrande  & MMLU & GSM8K & MBPP & HumanEval & Average \\ 
\midrule
Best Single Model       & 55.03\%          & 73.72\%          & 48.18\%          & 24.03\%          & 17.80\%          & 13.41\%          & 38.70\%          \\
\midrule
\multicolumn{8}{c}{\texttt{Which12}} \\
\midrule
Average     & 54.86\%          & 73.48\% & 49.42\%          & 32.98\%          & \textbf{23.60\%} & \textbf{21.34\%} & 42.61\% \\
Coefficient & 55.12\% & \textbf{73.64\%} & 50.13\% & 39.35\% & 21.80\% & \textbf{21.34\%} & \textbf{43.56\%} \\
Similarity$\uparrow$ & \textbf{56.48\%} & 73.32\% & \textbf{52.56\%} & 37.91\%          & 17.80\%          & 20.12\%          & 43.03\% \\
Similarity$\downarrow$ & 55.80\%          & 71.74\% & 52.39\%          & \textbf{47.99\%} & 16.40\%          & 15.85\%          & 43.36\%\\
\midrule
\multicolumn{8}{c}{\texttt{Which8}} \\
\midrule
Average     & \textbf{55.38\%} & \textbf{74.11\%} & 48.65\%          & 34.42\% & \textbf{25.20\%} & \textbf{23.17\%} & 43.49\%          \\
Coefficient & 55.12\%          & 73.64\%          & \textbf{50.13\%} & 39.35\% & 21.80\%          & 21.34\%          & \textbf{43.56\%} \\
Similarity$\uparrow$ & 54.95\% & 73.64\% & 49.00\% & 43.75\%          & 19.80\% & 11.59\% & 42.12\% \\
Similarity$\downarrow$ & 54.78\% & 72.30\% & 49.06\% & \textbf{47.23\%} & 21.20\% & 15.85\% & 43.40\% \\
\midrule
\multicolumn{8}{c}{\texttt{Which4}} \\
\midrule
Average     & 54.86\% & 73.16\% & 47.91\% & 37.00\%          & \textbf{24.00\%} & \textbf{21.34\%} & 43.05\%          \\
Coefficient & \textbf{55.12\%} & \textbf{74.03\%} & \textbf{48.18\%} & 41.93\% & 19.40\% & 14.63\% & 42.22\% \\
Similarity$\uparrow$ & 54.52\% & 73.24\% & 47.81\% & 41.77\%          & 21.20\%          & 20.73\%          & \textbf{43.21\%} \\
Similarity$\downarrow$ & 53.92\% & 73.56\% & 47.81\% & \textbf{48.45\%} & 18.20\%          & 10.98\%          & 42.15\%    \\
\bottomrule
\end{tabular}}
\end{table}

\begin{table}[h]
\centering
\caption{\small Comparison between different merging methods. } \label{tab:evo-methods}
\renewcommand\arraystretch{1}
\tabcolsep=0.1cm
\resizebox{0.8\linewidth}{!}{
\begin{tabular}{c|cccccc|c}
\toprule
\midrule
Merging Method & ARC  & WinoGrande  & MMLU & GSM8K & MBPP & HumanEval & Average \\ 
\midrule
\multicolumn{8}{c}{\texttt{Which12}} \\
\midrule
Linear               & \textbf{56.48\%}     & \textbf{73.56\%}     & 51.79\%              & 36.01\%              & \textbf{23.60\%}     & \textbf{20.12\%}     & \textbf{43.59\%}     \\
Task Arithmetic     & 51.54\%              & 69.14\%              & 51.07\%              & \textbf{54.66\%}     & 1.80\%               & 13.41\%              & 40.27\%              \\
DARE                & 51.19\%              & 70.09\%              & 51.03\%              & 53.53\%              & 6.80\%               & 12.80\%              & 40.91\%              \\
TIES                & 53.75\%              & 70.64\%              & \textbf{52.77\%}     & 49.36\%              & 17.20\%              & 2.44\%               & 41.03\%              \\
\midrule
\multicolumn{8}{c}{\texttt{Which8}} \\
\midrule
Linear              & \textbf{55.12\%}     & \textbf{73.64\%}     & \textbf{49.59\%}     & 40.64\%              & \textbf{22.40\%}     & 18.90\%              & 43.38\%              \\
Task Arithmetic     & 52.65\%              & 70.64\%              & 48.11\%              & 51.18\%              & 19.80\%              & \textbf{21.95\%}     & \textbf{44.05\%}     \\
DARE                & 52.56\%              & 71.19\%              & 49.00\%              & \textbf{53.37\%}     & 14.80\%              & 20.12\%              & 43.51\%              \\
TIES                & 50.26\%              & 71.27\%              & 48.58\%              & 47.69\%              & 18.40\%              & 0.61\%               & 39.47\%              \\
\midrule
\multicolumn{8}{c}{\texttt{Which4}} \\
\midrule
Linear              & \textbf{53.50\%}     & \textbf{73.01\%}     & \textbf{47.32\%}     & 45.79\%              & \textbf{20.20\%}     & 15.85\%              & 42.61\%              \\
Task Arithmetic     & 52.73\%              & 72.30\%              & 46.81\%              & \textbf{51.86\%}     & 18.20\%              & 18.29\%              & \textbf{43.36\%}     \\
DARE                & 51.45\%              & 71.67\%              & 45.61\%              & 51.55\%              & 16.60\%              & \textbf{20.12\%}     & 42.83\%              \\
TIES                & 50.51\%              & 71.98\%              & 46.62\%              & 49.43\%              & 16.40\%              & 1.22\%               & 39.36\%              \\
\bottomrule
\end{tabular}}
\end{table}

\begin{table}[h]
\centering
\caption{\small The impact of different group sizes on Evolutionary Strategy. } \label{tab:evo-group-size}
\renewcommand\arraystretch{1}
\tabcolsep=0.1cm
\resizebox{0.8\linewidth}{!}{
\begin{tabular}{c|cccccc|c}
\toprule
\midrule
Group Size & ARC  & WinoGrande  & MMLU & GSM8K & MBPP & HumanEval & Average \\ 
\midrule
\multicolumn{8}{c}{\texttt{Which12}} \\
\midrule
1  & 56.14\%          & 73.32\%          & 51.50\%          & 32.45\%          & \textbf{24.00\%} & \textbf{20.12\%} & 42.92\%          \\
4  & 56.31\%          & 73.72\%          & 52.04\%          & 33.43\%          & \textbf{24.00\%} & 18.90\%          & 43.07\%          \\
8  & \textbf{56.83\%} & \textbf{74.43\%} & \textbf{53.01\%} & \textbf{38.13\%} & 21.20\%          & 19.51\%          & \textbf{43.85\%} \\
32 & 56.48\%          & 73.56\%          & 51.79\%          & 36.01\%          & 23.60\%          & \textbf{20.12\%} & 43.59\%          \\
\midrule
\multicolumn{8}{c}{\texttt{Which8}} \\
\midrule
1  & 55.12\%          & 74.11\%          & 49.96\%          & 31.69\%          & \textbf{25.20\%} & 19.51\%          & 42.60\%          \\
4  & \textbf{56.06\%} & \textbf{74.66\%} & \textbf{50.04\%} & 33.59\%          & 24.20\%          & \textbf{21.95\%} & 43.42\%          \\
8  & 55.29\%          & 73.88\%          & 49.20\%          & 40.56\%          & 24.60\%          & 18.90\%          & \textbf{43.74\%} \\
32 & 55.12\%          & 73.64\%          & 49.59\%          & \textbf{40.64\%} & 22.40\%          & 18.90\%          & 43.38\%          \\
\midrule
\multicolumn{8}{c}{\texttt{Which4}} \\
\midrule
1  & \textbf{54.61\%} & 73.32\%          & \textbf{47.63\%} & 41.62\%          & 23.60\%          & 15.85\%          & 42.77\%          \\
4  & 52.90\%          & 73.32\%          & 46.99\%          & 43.06\%          & \textbf{24.00\%} & 20.73\%          & 43.50\%          \\
8  & 54.01\%          & \textbf{73.64\%} & 47.39\%          & 43.75\%          & 22.40\%          & \textbf{21.95\%} & \textbf{43.86\%} \\
32 & 53.50\%          & 73.01\%          & 47.32\%          & \textbf{45.79\%} & 20.20\%          & 15.85\%          & 42.61\%         \\
\midrule
\bottomrule
\end{tabular}}
\end{table}

\begin{table}[h]
\centering
\caption{\small More efficient merging strategy. } \label{tab:efficient-merge}
\renewcommand\arraystretch{1}
\tabcolsep=0.1cm
\resizebox{0.8\linewidth}{!}{
\begin{tabular}{c|ccccccc|c}
\toprule
\midrule
Strategy          & ARC     & WinoGrande & MMLU    & GSM8K   & MBPP    & HumanEval & Average & Round \\
\midrule
\multicolumn{8}{c}{\texttt{Which12}} \\
\midrule
Evo (Vanilla)     & $56.48\%$ & $73.56\%$    & $51.79\%$ & $36.01\%$ & $23.60\%$ & $20.12\%$   & $43.59\%$ & $200$   \\
Evo (Heuristic)   & $55.29\%$ & $72.85\%$    & $49.96\%$ & $40.56\%$ & $22.80\%$ & $18.29\%$   & $43.29\%$ & $127$   \\
\midrule
\multicolumn{8}{c}{\texttt{Which8}} \\
\midrule
Evo (Vanilla)     & $55.12\%$ & $73.64\%$    & $49.59\%$ & $40.64\%$ & $22.40\%$ & $18.90\%$   & $43.38\%$ & $200$   \\
Evo (Heuristic)   & $54.69\%$ & $72.93\%$    & $49.68\%$ & $45.19\%$ & $21.00\%$ & $19.51\%$   & $43.83\%$ & $71$    \\
\midrule
\multicolumn{8}{c}{\texttt{Which4}} \\
\midrule
Evo (Vanilla)     & $53.50\%$ & $73.01\%$    & $47.32\%$ & $45.79\%$ & $20.20\%$ & $15.85\%$   & $42.61\%$ & $200$   \\
Evo (Heuristic)   & $54.52\%$ & $73.56\%$    & $47.74\%$ & $40.71\%$ & $23.20\%$ & $21.95\%$   & $43.61\%$ & $69$    \\
\midrule
\bottomrule
\end{tabular}}
\end{table}

\subsection{Model Mixture}
For Model Level Mixture, we use more fine-grained prompts to construct the router, and report the results in Table~\ref{tab:q5-diverse-architecture-2}.
\begin{table}
\centering
\vspace{-18pt}
\caption{\small Better prompt vector for the mixture of Llama-2-7b-chat and CrystalChat. We highlight the better performance in \textbf{bold}.} \label{tab:q5-diverse-architecture-2}
\renewcommand\arraystretch{1}
\tabcolsep=0.1cm
\resizebox{0.8\linewidth}{!}{
\begin{tabular}{c|ccccccc}
\toprule
\midrule
Model & ARC  & WinoGrande  & MMLU & GSM8K & MBPP & HumanEval & Average \\ 
\midrule
Best Single Model & $52.05\%$ & $69.46\%$ & $50.77\%$ & $27.22\%$ & $\mathbf{39.60\%}$ & $\mathbf{35.98\%}$ & $45.85\%$ \\ 
\midrule
M-L-S  &  $\mathbf{51.88\%}$  &  $\mathbf{70.88\%}$  &  $\mathbf{52.44\%}$       & $\mathbf{32.52\%}$  &  $39.40\%$ & $31.10\%$  & $\mathbf{46.37\%}$   \\
\midrule
\bottomrule
\end{tabular}}
\vspace{-10pt}
\end{table}

\end{document}